\documentclass[11pt]{article}

\usepackage[margin=1in]{geometry}
\usepackage[OT1]{fontenc}
\usepackage{smile}
\usepackage{booktabs}
\allowdisplaybreaks
\usepackage{xcolor}
\usepackage{colortbl}
\definecolor{LightCyan}{rgb}{0.8,0.9,1}
\usepackage[colorlinks,
            linkcolor=red,
            anchorcolor=blue,
            citecolor=blue
            ]{hyperref}

\newcommand{\algname}{\texttt{VA-LSVI-UCB++}}
\newcommand{\algcell}[2]{\renewcommand{\arraystretch}{1.0}\begin{tabular}[c]{@{}l@{}}#1\\#2\end{tabular}}
\newcommand{\lsviucbpp}{\texttt{LSVI-UCB++}}
\newcommand{\lsviucb}{\texttt{LSVI-UCB}}
\newcommand{\ucrlvtrp}{\texttt{UCRL-VTR+}}
\newcommand{\hfucrlvtrp}{\texttt{HF-UCRL-VTR+}}
\newcommand{\ucrlave}{\texttt{UCRL-AVE}}
\newcommand{\ucbvi}{\texttt{UCBVI}}
\newcommand{\ucbadv}{\texttt{UCB-Advantage}}
\newcommand{\vara}{\texttt{VARA}}
\newcommand{\heavylsvi}{\texttt{HEAVY-LSVI-UCB}}

\newcommand{\Varkstar}{\mathrm{Var}_K^*}
\newcommand{\Varkpi}{\mathrm{Var}_K^\pi}
\newcommand{\Vest}{V_{\mathrm{est}}}
\newcommand{\Vtotalstar}{V_{\mathrm{total}}^*}
\newcommand{\Econc}{\cE_{\mathrm{conc}}}
\newcommand{\EFreed}{\cE_{\mathrm{Freed}}}
\newcommand{\klast}{k_{\mathrm{last}}}
\newcommand{\lmax}{l_{\max}}
\newcommand{\ul}{\mathrm{ul}}

\title{\huge Efficient Variance-Aware Reinforcement Learning in Time-Inhomogeneous Linear MDPs}

\author{
  Qingyue Zhao\thanks{Equal contribution} \thanks{Department of Computer Science, University of California, Los Angeles, CA 90095, USA; e-mail: \texttt{zhaoqy24@cs.ucla.edu}}
  \qquad
  Qiwei Di\footnotemark[1] \thanks{Department of Computer Science, University of California, Los Angeles, CA 90095, USA; e-mail: \texttt{qiwei2000@cs.ucla.edu}}
  \qquad
  Jiafan He\thanks{Department of Computer Science, University of California, Los Angeles, CA 90095, USA; e-mail: \texttt{jiafanhe19@ucla.edu}}
  \qquad
  Heyang Zhao\thanks{Department of Computer Science, University of California, Los Angeles, CA 90095, USA; e-mail: \texttt{hyzhao@cs.ucla.edu}}
  \\[0.5em]
  Kaixuan Ji\thanks{Department of Computer Science, University of California, Los Angeles, CA 90095, USA; e-mail: \texttt{kaixuanji@cs.ucla.edu}}
  \qquad
  Quanquan Gu\thanks{Department of Computer Science, University of California, Los Angeles, CA 90095, USA; e-mail: \texttt{qgu@cs.ucla.edu}}
}
\date{}

\begin{document}

\maketitle

\begin{abstract}
We study variance-dependent reinforcement learning in episodic time-inhomogeneous linear Markov decision processes (MDPs) with feature dimension $d$, horizon $H$, and $K$ episodes. While recent works have established variance-dependent regret with $\tilde{\cO}(1)$ deterministic-regime overhead for linear mixture MDPs \citep{zhao2023variance} and worst-case nearly minimax optimal regret $\tilde{\cO}(d\sqrt{H^3 K} + d^7 H^8)$ for linear MDPs \citep{he2023nearly}, existing variance-aware analyses for linear MDPs \citep{li2023variance,huang2023tackling} rely on a coarse concentration step whose resulting regret guarantees retain a polynomial $\tilde{\cO}(K^{1/4})$ additive term even in deterministic environments. We propose \algname{}, a computationally efficient algorithm that achieves the variance-dependent regret bound
\[
  \Regret(K) = \tilde{\cO}\!\rbr{d\sqrt{H \cdot \min\cbr{\Varkstar, \Varkpi}} + d^5 H^4},
\]
where $\Varkstar$ and $\Varkpi$ denote the cumulative conditional variances of the optimal value function and the played policies' value functions along the realized trajectory. Our result adapts to the minimum of the two realized trajectory variances, recovers the worst-case minimax optimal leading term $\tilde{\cO}(d\sqrt{H^3 K})$, and collapses to $\tilde{\cO}(d^5 H^4)$ in deterministic environments, free of polynomial-in-$K$ dependency. At the core of our analysis is a simultaneous three-way reference-advantage decomposition for the optimistic, pessimistic, and squared value function estimators coupled with variance self-bounding and Freedman-type value-difference arguments, which tightens the pessimistic and second-moment confidence radii by a factor of $\tilde{\cO}(d H)$ and removes the inverse-variance-floor blowup.
\end{abstract}

\section{Introduction}

Designing statistically and computationally efficient algorithms for reinforcement learning (RL) with function approximation is a central problem in modern statistical learning theory. To handle large or continuous state spaces, a major line of research studies Markov decision processes (MDPs) equipped with linear structure, most notably \emph{linear mixture MDPs} \citep{modi2020sample,jia2020model,ayoub2020model,zhou2021nearly,zhou2022computationally,zhao2023variance}, where the transition kernel is a linear combination of $d$ basis transition models $\PP_h(s'\mid s,a) = \inner{\bphi(s'\mid s,a)}{\btheta_h^*}$, and \emph{linear MDPs} \citep{yang2019sample,jin2020provably,hu2022nearly,he2023nearly,agarwal2023voql}, where both the transition kernel $\PP_h(s'\mid s,a) = \inner{\bphi(s,a)}{\dd\btheta_h(s')}$ and the reward function are linear in a known $d$-dimensional state-action feature map $\bphi(s,a)$ with unknown signed measures $\btheta_h(\cdot)$ over the state space $\cS$.

While worst-case minimax regret bounds quantify algorithm performance in the hardest environment instances, practical RL environments often exhibit low or even zero transition stochasticity. This observation has motivated a surge of interest in \emph{variance-dependent} (or second-order) regret bounds \citep{zanette2019tighter,zhou2021nearly,zhang2021improved,kim2022improved,zhou2022computationally,zhao2023variance,li2023variance,huang2023tackling,zhou2023sharp,foster2024behavior}, which scale with the transition variance rather than the worst-case horizon bound and ideally interpolate between the $\tilde{\cO}(\sqrt{K})$ minimax rate in stochastic regimes and $\tilde{\cO}(1)$ (polylogarithmic in $K$) regret in deterministic regimes. For time-homogeneous linear mixture MDPs under $\sum_{h=1}^H r_h \le 1$, \citet{zhao2023variance} proposed \ucrlave{}, achieving a variance-dependent regret of $\tilde{\cO}(d\sqrt{\Varkstar} + d^2)$ without prior knowledge of the transition variances.

From a statistical perspective, however, linear MDPs are substantially more challenging to analyze than linear mixture MDPs. In linear mixture MDPs, for any next-stage value function $V_{k,h+1}$, the integrated feature vector $\bphi_{V_{k,h+1}}(s,a) = \sum_{s'\in\cS}\bphi(s'\mid s,a)V_{k,h+1}(s')$ is explicitly computable by the learner, and the regression target parameter $\btheta_h^* \in \RR^d$ is fixed and independent of $V_{k,h+1}$. Exploiting this fixed-target structure, \citet{zhao2023variance} directly applied self-normalized martingale concentration to the single target vector $\btheta_h^*$ without requiring an $\epsilon$-net covering over the value function class. In contrast, in linear MDPs, the feature vector $\bphi(s_h^i, a_h^i)$ is fixed while the true Bellman backup parameter $\wb_h(V_{k,h+1}) = \int_{\cS} V_{k,h+1}(s')\dd\btheta_h(s')$ depends on the random, data-dependent value function $V_{k,h+1}$ constructed from later stages. This statistical dependency breaks martingale adaptedness and necessitates a uniform convergence ($\epsilon$-net covering) argument over the value function class \citep{jin2020provably}. Because the optimistic value function class has log-covering number $\tilde{\cO}(d^2)$ (or $\tilde{\cO}(d^3 H)$ under rare-switching updates), naive uniform convergence inflates the confidence radius by $\sqrt{d}$, leading to the suboptimal $\tilde{\cO}(\sqrt{d^3 H^4 K})$ regret of \lsviucb{} \citep{jin2020provably}.

Recently, \citet{he2023nearly} (concurrently with \citet{agarwal2023voql}) resolved the worst-case minimax optimality for linear MDPs by proposing \lsviucbpp{}, which attains $\tilde{\cO}(d\sqrt{H^3 K} + d^7 H^8)$ regret via variance-weighted ridge regression, reference-advantage decomposition $V_{k,h+1} = V_{h+1}^* + (V_{k,h+1} - V_{h+1}^*)$, pointwise monotonic value updates, and rare-switching policy updates. Yet, despite employing variance-weighted regression, the guarantee of \lsviucbpp{} is purely worst-case and does not exploit low or zero transition variance: even in a deterministic linear MDP where all transition variances vanish, the regret bound of \citet{he2023nearly} remains $\tilde{\cO}(d\sqrt{H^3 K} + d^7 H^8)$. Specifically, the analysis of \citet{he2023nearly} relies on (i) a coarse, variance-oblivious concentration bound over the full function classes with initial confidence radius scaling inversely with the variance floor $\sigma_{\min}$ (with $\sigma_{\min} = \sqrt{H}$ used in the coarse analysis, yielding large radii $\bar{\beta} = \tilde{\cO}(\sqrt{d^3 H^2})$ and $\tilde{\beta} = \tilde{\cO}(\sqrt{d^3 H^4})$ for the pessimistic and second-moment estimators), and (ii) Azuma--Hoeffding inequalities on the transition martingale differences of $V_{k,h+1} - V_{h+1}^{\pi^k}$ and $V_{k,h+1} - \check{V}_{k,h+1}$ that inject unconditional $\sqrt{H^3 K}$ and $\sqrt{H^5 K}$ terms. Subsequent works by \citet{li2023variance} (\vara{}) and \citet{huang2023tackling} (\heavylsvi{}) replaced the Azuma--Hoeffding step with Freedman-type arguments to obtain variance-aware leading terms $\tilde{\cO}(d\sqrt{H K \cG^*})$ and $\tilde{\cO}(d\sqrt{H K \cV^*})$ (where $\cG^*, \cV^*$ denote expected occupancy-measure variances), but both still retained the coarse first-stage confidence radius $\beta_0 = \tilde{\cO}(H\sqrt{d^3 H}/\sigma_{\min})$ (and $H\beta_0 = \tilde{\cO}(H^2\sqrt{d^3 H}/\sigma_{\min})$ for the second-moment target); trading off $\sigma_{\min} \propto K^{-1/4}$ in their regret upper bounds leaves an unconditional $\tilde{\cO}(K^{1/4})$ polynomial additive term even in deterministic MDPs. Thus, a natural question remains open:
\begin{center}
\emph{Can we design a computationally efficient algorithm for linear MDPs that achieves variance-dependent regret adapting to the realized trajectory variance $\min\{\Varkstar, \Varkpi\}$ with a $K$-independent (up to polylogarithmic factors) lower-order remainder?}
\end{center}

\begin{table}[t]
\centering
\caption{Comparison of regret bounds for episodic time-inhomogeneous RL with linear function approximation under per-step reward $r_h \in [0,1]$. Regret upper bounds are shown up to universal constants and polylogarithmic factors. Here $d$ is the feature dimension, $H$ is the episode horizon, $K$ is the number of episodes, $\Vtotalstar = \sum_{k=1}^K V_1^*(s_1^k)$ is the cumulative optimal value, $\cG^*$ and $\cV^*$ are expected occupancy-measure variances \citep{li2023variance,huang2023tackling}, and $\Varkstar, \Varkpi$ are the realized trajectory-dependent cumulative variances defined in \eqref{eq:def-traj-var}. For \vara{} \citep[Theorem~4.1]{li2023variance} and \heavylsvi{} \citep[Theorem~F.13]{huang2023tackling}, bounds are shown adapted to known deterministic rewards $r_h \in [0,1]$ ($\cH = H$, zero reward noise) and optimized over $\sigma_{\min}$. For time-homogeneous horizon-free results under $\sum_{h=1}^H r_h \le 1$, see Table~\ref{tab:comparison-homog} in Appendix~\ref{app:homog-comparison}.}
\label{tab:comparison}
\vspace{4pt}
\begingroup
\footnotesize
\setlength{\tabcolsep}{2.5pt}
\renewcommand{\arraystretch}{1.35}
\begin{tabular}{l c c c}
\toprule
\textbf{Algorithm} & \textbf{Regret Bound} & \textbf{Deterministic Regime} & \textbf{Efficient?} \\
\midrule
\multicolumn{4}{l}{\textit{Time-Inhomogeneous Linear Mixture MDPs}} \\
\algcell{\ucrlvtrp{}}{\citep{zhou2021nearly}}
& $d\sqrt{H^3 K}+\sqrt{d H^4 K}+d^2 H^3+d^3 H^2$
& $d\sqrt{H^3 K}+\sqrt{d H^4 K}+d^2 H^3+d^3 H^2$
& $\checkmark$ \\
\midrule
\multicolumn{4}{l}{\textit{Time-Inhomogeneous Linear MDPs}} \\
\algcell{\lsviucb{}}{\citep{jin2020provably}}
& $\sqrt{d^3 H^4 K}$
& $\sqrt{d^3 H^4 K}$
& $\checkmark$ \\
\addlinespace[2pt]
\algcell{\texttt{ELEANOR}}{\citep{zanette2020learning}}
& $d\sqrt{H^4 K}$
& $d\sqrt{H^4 K}$
& $\times$ \\
\addlinespace[2pt]
\algcell{\texttt{Force}}{\citep{wagenmaker2022first}}
& $\sqrt{d^3 H^3\Vtotalstar}+d^{3.5}H^3$
& $\sqrt{d^3 H^3\Vtotalstar}+d^{3.5}H^3$
& $\times$ \\
\addlinespace[2pt]
\algcell{\lsviucbpp{}}{\citep{he2023nearly}}
& $d\sqrt{H^3 K}+d^7 H^8$
& $d\sqrt{H^3 K}+d^7 H^8$
& $\checkmark$ \\
\addlinespace[2pt]
\algcell{\vara{}}{\citep{li2023variance}}
& $d\sqrt{H K\cG^*}+d^{3.5}H^{2.75}K^{1/4}+d^5 H^4$
& $d^{3.5}H^{2.75}K^{1/4}+d^5 H^4$
& $\checkmark$ \\
\addlinespace[2pt]
\algcell{\heavylsvi{}}{\citep{huang2023tackling}}
& $d\sqrt{H K\cV^*}+d^{3.5}H^{2.75}K^{1/4}+d^5 H^4$
& $d^{3.5}H^{2.75}K^{1/4}+d^5 H^4$
& $\checkmark$ \\
\addlinespace[2pt]
\rowcolor{LightCyan}
\algcell{\textbf{\algname{}}}{\textbf{(Theorem~\ref{thm:main})}}
& {\boldmath$d\sqrt{H\cdot\min\{\Varkstar,\Varkpi\}}+d^5 H^4$}
& $\bm{d^5 H^4}$
& $\bm{\checkmark}$ \\
\bottomrule
\end{tabular}
\endgroup
\end{table}

\paragraph{Main contributions.}
In this paper, we answer the open question above affirmatively by proposing \algname{} (Algorithm~\ref{alg:va-lsvi-ucbpp}), a computationally efficient algorithm for episodic time-inhomogeneous linear MDPs. A detailed comparison with prior results for time-inhomogeneous MDPs is provided in Table~\ref{tab:comparison} (with a companion comparison for time-homogeneous horizon-free RL in Appendix~\ref{app:homog-comparison} and Table~\ref{tab:comparison-homog}). Our main contributions are summarized as follows:

\begin{itemize}
  \item \textbf{Variance-dependent regret with worst-case minimax optimal leading term.} We prove that for any episodic linear MDP with $r_h \in [0,1]$, \algname{} achieves high-probability regret
  \[
    \Regret(K) = \tilde{\cO}\!\rbr{d\sqrt{H \cdot \min\cbr{\Varkstar, \Varkpi}} + d^5 H^4},
  \]
  while performing only $\cO(d H \log K)$ policy updates and running in $\tilde{\cO}(d^4 H^3 |\cA| K)$ total time (Theorem~\ref{thm:main}). Here $\Varkstar = \sum_{k,h} [\VV_h V_{h+1}^*](s_h^k, a_h^k)$ and $\Varkpi = \sum_{k,h} [\VV_h V_{h+1}^{\pi^k}](s_h^k, a_h^k)$ denote the realized trajectory-dependent cumulative variances along the learner's state-action trajectory (defined in \eqref{eq:def-traj-var}). Our bound adapts to $\min\{\Varkstar, \Varkpi\}$, recovering the worst-case minimax optimal leading term $\tilde{\cO}(d\sqrt{H^3 K})$ that matches the $\Omega(d\sqrt{H^3 K})$ lower bound of \citet{zhou2021nearly} and \citet{he2023nearly} up to logarithmic factors. In deterministic MDPs where $\Varkstar = \Varkpi = 0$, our regret bound collapses to $\tilde{\cO}(d^5 H^4)$, eliminating the $\tilde{\cO}(K^{1/4})$ polynomial additive term of \citet{li2023variance} and \citet{huang2023tackling} while improving the burn-in overhead of \lsviucbpp{} \citep{he2023nearly} from $\tilde{\cO}(d^7 H^8)$ to $\tilde{\cO}(d^5 H^4)$.

  \item \textbf{Three-way reference-advantage decomposition and double induction.} To eliminate the coarse Hoeffding concentration step in \citet{he2023nearly} (which required a constant floor $\sigma_{k,h}^2 \ge H$ and inflated the pessimistic and second-moment confidence radii to $\bar{\beta} = \tilde{\cO}(\sqrt{d^3 H^2})$ and $\tilde{\beta} = \tilde{\cO}(\sqrt{d^3 H^4})$), we apply reference-advantage decompositions to all three regression targets simultaneously:
  \begin{align*}
    V_{k,h+1} &= V_{h+1}^* + \big(V_{k,h+1} - V_{h+1}^*\big), \qquad \check{V}_{k,h+1} = V_{h+1}^* - \big(V_{h+1}^* - \check{V}_{k,h+1}\big), \\
    V_{k,h+1}^2 &= (V_{h+1}^*)^2 + \big(V_{k,h+1}^2 - (V_{h+1}^*)^2\big).
  \end{align*}
  Combining the conditional variance bound $[\VV_h (V_{h+1}^*)^2] \le 4 H^2 [\VV_h V_{h+1}^*]$ \citep{chen2021implicit,chen2022improved,foster2024behavior} with the pointwise monotonicity of $V_{k,h+1}$ and $\check{V}_{k,h+1}$ \citep{azar2017minimax,he2023nearly}, we establish variance-normalized Bernstein concentration for all three estimators on a single unconditional high-probability event $\Econc$ via a deterministic double induction over $h = H \to 1$ and $k = 1 \to K$ (Lemma~\ref{lem:double-induction}). This shaves a factor of $\tilde{\cO}(d H)$ off both $\bar{\beta}$ and $\tilde{\beta}$, reducing them to $\tilde{\cO}(\sqrt{d})$ and $\tilde{\cO}(H\sqrt{d})$, and allows setting the variance floor to $\alpha^2 = H^2/K$.

  \item \textbf{Freedman-type analysis for value differences and trajectory variance.} Because the optimistic-to-played gap $V_{k,h+1} - V_{h+1}^{\pi^k} \in [0, H]$ and the optimistic-to-pessimistic gap $V_{k,h+1} - \check{V}_{k,h+1} \in [0, H]$ are non-negative under optimism and pessimism, their conditional variances are self-bounded by $H[\PP_h(V_{k,h+1} - V_{h+1}^{\pi^k})]$ and $H[\PP_h(V_{k,h+1} - \check{V}_{k,h+1})]$. By applying a time-uniform Freedman inequality and absorbing the conditional variance sum into the left-hand side (Lemma~\ref{lem:freedman-value-diff}), we completely eliminate the unconditional $\sqrt{H^3 K}$ and $\sqrt{H^5 K}$ Azuma--Hoeffding errors in \citet{he2023nearly}. Using a related variance self-bounding argument in the spirit of \citet{chen2021implicit,chen2022improved} and \citet{foster2024behavior}, we also establish an immediate first-order regret bound of $\tilde{\cO}(d\sqrt{H^2 \Vtotalstar} + d^5 H^4)$ (Corollary~\ref{cor:first-order}).
\end{itemize}

\paragraph{Notation.}
We use lowercase letters to denote scalars, and boldface lowercase and uppercase letters to denote vectors and matrices, respectively. For a positive integer $n \in \NN_+$, we denote by $[n]$ the set $\{1, \dots, n\}$. For a vector $\xb \in \RR^d$ and a positive semi-definite matrix $\bSigma \in \RR^{d\times d}$, we denote by $\|\xb\|_2$ the Euclidean norm and define $\|\xb\|_{\bSigma} := \sqrt{\xb^\top \bSigma \xb}$. For a matrix $\bGamma \in \RR^{d\times d}$, $\|\bGamma\|_F$ denotes its Frobenius norm, and $\Ab \succeq \Bb$ (resp.\ $\Ab \preceq \Bb$) means that $\Ab - \Bb$ is positive (resp.\ negative) semi-definite. For any real function $f: \cS \to \RR$, we define $\|f\|_\infty := \sup_{s\in\cS}|f(s)|$. We use $\ind\{\cdot\}$ to denote the indicator function. For $x \in \RR$ and $a \le b$, we define the positive-part operator $(x)_+ := \max\{x, 0\}$ and the truncation operator $[x]_{[a,b]} := \min\{\max\{x, a\}, b\}$. For two positive sequences $\{a_n\}$ and $\{b_n\}$, we write $a_n = \cO(b_n)$ (resp.\ $a_n = \Omega(b_n)$) if there exists an absolute constant $C > 0$ such that $a_n \le C b_n$ (resp.\ $a_n \ge C b_n$), and use $\tilde{\cO}(\cdot)$ and $\tilde{\Omega}(\cdot)$ to hide polylogarithmic factors in $d, H, K, 1/\delta$.

\section{Related Work}

\paragraph{RL with linear function approximation.}
A rich body of literature has studied provably efficient RL with linear function approximation under various structural assumptions, including low Bellman rank \citep{jiang2017contextual,dann2018oracle}, low inherent Bellman error \citep{zanette2020learning}, linear mixture MDPs \citep{modi2020sample,jia2020model,ayoub2020model,zhou2021provably,zhou2021nearly,zhou2022computationally,zhao2023variance}, and linear MDPs \citep{yang2019sample,jin2020provably,zanette2020frequentist,he2021logarithmic,wang2021provably,wagenmaker2022first,hu2022nearly,he2023nearly,agarwal2023voql,li2023variance,huang2023tackling}. For episodic linear MDPs, \citet{jin2020provably} proposed \lsviucb{} and established an $\tilde{\cO}(\sqrt{d^3 H^4 K})$ regret bound. To close the $\sqrt{d H}$ gap to the $\Omega(d\sqrt{H^3 K})$ lower bound \citep{zhou2021nearly}, \citet{hu2022nearly} attempted to combine reference-advantage decomposition with an over-optimistic value function sequence, which was subsequently found to contain a technical flaw in its covering argument \citep[Appendix~A]{he2023nearly}. \citet{he2023nearly} and \citet{agarwal2023voql} independently achieved the worst-case minimax optimal $\tilde{\cO}(d\sqrt{H^3 K})$ leading rate via \lsviucbpp{} and \texttt{VOQL}, respectively. Building on \lsviucbpp{}, \citet{li2023variance} (\vara{}) and \citet{huang2023tackling} (\heavylsvi{}) studied variance- and central-moment-aware regret in linear MDPs with heavy-tailed rewards; however, because their transition analyses still rely on a coarse concentration stage whose radius scales inversely with the variance floor $\sigma_{\min}$, trading off $\sigma_{\min} \propto K^{-1/4}$ introduces an unconditional $\tilde{\cO}(K^{1/4})$ polynomial regret overhead even in deterministic MDPs.

\paragraph{Variance-aware bandits and problem-dependent RL.}
Exploiting empirical variance via Bernstein- or Freedman-type concentration inequalities has been pivotal for shaving horizon factors in tabular RL. Pioneered by \citet{lattimore2012pac} and \citet{azar2013minimax,azar2017minimax}, empirical Bernstein bonuses combined with the Law of Total Variance enabled \ucbvi{} \citep{azar2017minimax} and \ucbadv{} \citep{zhang2020almost} to achieve minimax optimal $\tilde{\cO}(\sqrt{H^2 S A K})$ and $\tilde{\cO}(\sqrt{H^3 S A K})$ regret in time-homogeneous and time-inhomogeneous tabular MDPs, respectively, and subsequently enabled horizon-free regret bounds under $\sum_{h=1}^H r_h \le 1$ \citep{wang2020long,zhang2021reinforcement,li2022settling,zhang2022horizon}. In parallel, self-bounding Freedman-type arguments coupled with squared-value variance bounds ($\Var(X^2) \le 4\|X\|_\infty^2 \Var(X)$) have played a central role in controlling trajectory variances in stochastic shortest path (SSP) problems \citep{chen2021implicit,chen2022improved}, problem-dependent tabular RL \citep{zanette2019tighter,zhou2023sharp}, and horizon-independent imitation learning \citep{foster2024behavior}.

In linear bandits with heteroscedastic noise of variance $\sigma_k^2$, \citet{kirschner2018information} and \citet{zhou2021nearly} studied known-variance weighted ridge regression, and \citet{zhou2022computationally} introduced variance- and uncertainty-aware weights to achieve $\tilde{\cO}(d\sqrt{\sum_{k=1}^K \sigma_k^2} + d)$ regret (a weighting mechanism also leveraged in corruption-robust linear bandits by \citet{he2022nearly}). For unknown noise variances, \citet{zhang2021improved} and \citet{kim2022improved} developed statistically near-optimal but computationally inefficient algorithms (\texttt{VOFUL}/\texttt{VOFUL2}) over non-convex second-order confidence sets, which they extended to horizon-free linear mixture MDPs (\texttt{VARLin}/\texttt{VARLin2}). \citet{zhou2022computationally} achieved the first computationally efficient horizon-free $\tilde{\cO}(d\sqrt{K} + d^2)$ regret in time-homogeneous linear mixture MDPs via higher-order moment estimators (\hfucrlvtrp{}), and \citet{zhao2023variance} further achieved variance-dependent $\tilde{\cO}(d\sqrt{\Varkstar} + d^2)$ regret via multi-layer stratification (\ucrlave{}). For linear MDPs, \citet{wagenmaker2022first} obtained a computationally inefficient first-order bound $\tilde{\cO}(\sqrt{d^3 H^3 \Vtotalstar} + d^{3.5}H^3)$, \citet{chen2022improved} designed an inefficient global-optimization algorithm for linear SSP, \citet{zhang2024horizon} established a computationally inefficient worst-case horizon-free bound $\tilde{\cO}(d^{5.5}\sqrt{K} + d^{6.5})$ for time-homogeneous linear MDPs under $\sum_{h=1}^H r_h \le 1$ (see Appendix~\ref{app:homog-comparison}), and \citet{li2023variance} and \citet{huang2023tackling} obtained variance- and first-order bounds for time-inhomogeneous linear MDPs with $\tilde{\cO}(K^{1/4})$ lower-order overhead. Our \algname{} algorithm simultaneously achieves computational efficiency, adaptation to the realized trajectory variance $\min\{\Varkstar, \Varkpi\}$, and a lower-order remainder $\tilde{\cO}(d^5 H^4)$ free of polynomial-in-$K$ dependency.

\section{Preliminaries}

\paragraph{Episodic time-inhomogeneous MDPs.}
We consider an episodic time-inhomogeneous MDP denoted by a tuple $\cM = (\cS, \cA, H, \{\PP_h\}_{h=1}^H, \{r_h\}_{h=1}^H)$, where $\cS$ is a measurable state space (possibly infinite), $\cA$ is a finite action space, $H \in \NN_+$ is the episode horizon, $\PP_h(\cdot \mid s,a)$ is the transition probability kernel at stage $h \in [H]$, and $r_h: \cS \times \cA \to [0,1]$ is the deterministic reward function at stage $h \in [H]$.\footnote{Following standard convention \citep{jin2020provably,he2023nearly}, we assume deterministic known rewards to isolate the statistical complexity of learning the transition kernels.} A deterministic Markov policy $\pi = \{\pi_h\}_{h=1}^H$ is a sequence of $H$ mappings $\pi_h: \cS \to \cA$.

For any policy $\pi$ and stage $h \in [H]$, the state value function $V_h^\pi: \cS \to [0,H]$ and action-value function $Q_h^\pi: \cS \times \cA \to [0,H]$ are defined as the expected cumulative rewards starting from stage $h$:
\[
  Q_h^\pi(s,a) := r_h(s,a) + \EE\!\sbr{\sum_{h'=h+1}^H r_{h'}\big(s_{h'}, \pi_{h'}(s_{h'})\big) \,\middle|\, s_h = s, a_h = a}, \qquad V_h^\pi(s) := Q_h^\pi\big(s, \pi_h(s)\big),
\]
with boundary condition $V_{H+1}^\pi(s) := 0$ for all $s \in \cS$. For any measurable function $V: \cS \to \RR$, we define the one-step conditional expectation operator $\PP_h$ and conditional variance operator $\VV_h$ at stage $h \in [H]$ by
\[
  [\PP_h V](s,a) := \EE_{s' \sim \PP_h(\cdot \mid s,a)}\big[V(s')\big], \qquad [\VV_h V](s,a) := [\PP_h V^2](s,a) - \big([\PP_h V](s,a)\big)^2,
\]
where $V^2$ denotes the pointwise squared function $s \mapsto (V(s))^2$. The optimal action-value function $Q_h^*(s,a) := \sup_\pi Q_h^\pi(s,a)$ and optimal value function $V_h^*(s) := \sup_\pi V_h^\pi(s)$ satisfy the Bellman equations:
\[
  Q_h^\pi(s,a) = r_h(s,a) + [\PP_h V_{h+1}^\pi](s,a), \quad Q_h^*(s,a) = r_h(s,a) + [\PP_h V_{h+1}^*](s,a), \quad V_h^*(s) = \max_{a\in\cA} Q_h^*(s,a),
\]
with $V_{H+1}^*(s) = 0$ for all $s \in \cS$.

\paragraph{Linear MDPs.}
We focus on linear MDPs \citep{yang2019sample,jin2020provably}, defined as follows.

\begin{definition}[Linear MDP, \citealt{jin2020provably}]
\label{def:linear-mdp}
An MDP $\cM = (\cS, \cA, H, \{\PP_h\}_{h=1}^H, \{r_h\}_{h=1}^H)$ is a linear MDP with a known feature map $\bphi: \cS \times \cA \to \RR^d$ if for every stage $h \in [H]$, there exist $d$ unknown signed measures $\btheta_h = (\theta_{h,1}, \dots, \theta_{h,d})^\top$ over $\cS$ and an unknown vector $\bmu_h \in \RR^d$ such that for all $(s,a,s') \in \cS \times \cA \times \cS$,
\[
  \PP_h(s' \mid s,a) = \inner{\bphi(s,a)}{\dd\btheta_h(s')}, \qquad r_h(s,a) = \inner{\bphi(s,a)}{\bmu_h}.
\]
We assume that $\|\bphi(s,a)\|_2 \le 1$ for all $(s,a) \in \cS \times \cA$, $\|\bmu_h\|_2 \le \sqrt{d}$, and use the following bounded-integration normalization for the signed transition measures (cf.\ \citealp{jin2020provably,he2023nearly}):
\[
  \|\btheta_h(\cS)\|_2 := \sup_{f: \cS \to [-1,1]} \nbr{\int_{\cS} f(s')\dd\btheta_h(s')}_2 \le \sqrt{d},
\]
which holds in particular whenever the coordinate total variations satisfy $\sum_{j=1}^d (|\theta_{h,j}|(\cS))^2 \le d$ (encompassing tabular MDPs with $d = |\cS||\cA|$ and simplex feature spaces; see Examples~2.1--2.2 of \citealt{jin2020provably}).
\end{definition}

A foundational property of Definition~\ref{def:linear-mdp} is that for any bounded function $V: \cS \to \RR$, both the first moment $[\PP_h V](s,a)$ and the second moment $[\PP_h V^2](s,a)$ are linear functions of the feature map $\bphi(s,a)$.

\begin{proposition}[Linear Representation of Moments, \citealt{jin2020provably}]
\label{prop:linear-moments}
Under Definition~\ref{def:linear-mdp}, for any stage $h \in [H]$ and any measurable function $f: \cS \to \RR$ satisfying $\|f\|_\infty \le B$, the integrated parameter vector $\wb_h(f) := \int_{\cS} f(s')\dd\btheta_h(s') \in \RR^d$ satisfies $\|\wb_h(f)\|_2 \le B\sqrt{d}$ and
\[
  [\PP_h f](s,a) = \inner{\wb_h(f)}{\bphi(s,a)} \qquad \forall (s,a) \in \cS \times \cA.
\]
In particular, for any value function $V: \cS \to [0,H]$, we have $[\PP_h V](s,a) = \inner{\wb_h(V)}{\bphi(s,a)}$ with $\|\wb_h(V)\|_2 \le H\sqrt{d}$, and $[\PP_h V^2](s,a) = \inner{\wb_h(V^2)}{\bphi(s,a)}$ with $\|\wb_h(V^2)\|_2 \le H^2\sqrt{d}$.
\end{proposition}

\paragraph{Learning protocol, regret, and trajectory variances.}
The learner interacts with the environment for $K$ episodes. At the beginning of each episode $k \in [K]$, the environment selects an arbitrary (possibly history-dependent) initial state $s_1^k \in \cS$, and the learner chooses a policy $\pi^k = \{\pi_h^k\}_{h=1}^H$ based on historical observations. Executing $\pi^k$ generates a trajectory $(s_1^k, a_1^k, s_2^k, a_2^k, \dots, s_H^k, a_H^k, s_{H+1}^k)$ where $a_h^k = \pi_h^k(s_h^k)$ and $s_{h+1}^k \sim \PP_h(\cdot \mid s_h^k, a_h^k)$. We denote by $\cF_{k,h}$ the $\sigma$-algebra generated by all random variables up to and including $(s_h^k, a_h^k)$ at stage $h$ of episode $k$ prior to observing $s_{h+1}^k$.

The performance of the learner over $K$ episodes is measured by the cumulative regret:
\[
  \Regret(K) := \sum_{k=1}^K \rbr{V_1^*(s_1^k) - V_1^{\pi^k}(s_1^k)}.
\]
To characterize the transition stochasticity encountered during learning, we define the cumulative trajectory variances of the optimal value function and of the played policies' value functions as
\begin{equation}
  \Varkstar := \sum_{k=1}^K \sum_{h=1}^H [\VV_h V_{h+1}^*](s_h^k, a_h^k), \qquad \Varkpi := \sum_{k=1}^K \sum_{h=1}^H [\VV_h V_{h+1}^{\pi^k}](s_h^k, a_h^k). \label{eq:def-traj-var}
\end{equation}

Unlike population variance notions defined by taking expectations under a fixed policy's rollout distribution \citep{zanette2019tighter,li2023variance,huang2023tackling,foster2024behavior}, $\Varkstar$ \citep{zhao2023variance,zhou2023sharp} and $\Varkpi$ in \eqref{eq:def-traj-var} are realized, trajectory-dependent random variables evaluated along the learner's adaptive trajectory $\{(s_h^k, a_h^k)\}_{k\in[K], h\in[H]}$. In particular, $\Varkstar = \Varkpi = 0$ almost surely in deterministic MDPs, while in general stochastic MDPs $\Varkpi$ is bounded with high probability by both $\cO(H^2(K + \ln(2/\delta)))$ and $\cO(H\Vtotalstar + H^2\ln(4/\delta))$ (Lemma~\ref{lem:traj-var}).

\section{The Proposed Algorithm}

In this section, we present our variance-aware algorithm \algname{} in Algorithm~\ref{alg:va-lsvi-ucbpp} and describe its core algorithmic components.

\begin{algorithm}[!t]
\caption{\algname{}: Variance-Aware \lsviucbpp{} for Linear MDPs}
\label{alg:va-lsvi-ucbpp}
\begin{algorithmic}[1]
\REQUIRE Failure probability $\delta \in (0,1)$, regularization $\lambda = 1/H^2$, variance floor $\alpha = H/\sqrt{K}$, uncertainty weight $\gamma = 2 d^{1.5} H^{1.5} \iota$, confidence radii $\beta, \bar{\beta}, \tilde{\beta}$ defined in \eqref{eq:def-radii}.
\STATE Initialize $\bSigma_{1,h} \leftarrow \lambda \Ib$, $Q_{0,h}(s,a) \leftarrow H$, $\check{Q}_{0,h}(s,a) \leftarrow 0$, $V_{k,H+1}(s) \leftarrow 0$, $\check{V}_{k,H+1}(s) \leftarrow 0$ for all $(s,a,h,k) \in \cS \times \cA \times [H] \times [K]$, and $\klast \leftarrow 1$.
\FOR{episode $k = 1, 2, \dots, K$}
  \STATE Observe initial state $s_1^k$.
  \IF{there exists $h' \in [H]$ such that $\det(\bSigma_{k,h'}) \ge 2 \det(\bSigma_{\klast,h'})$ or $k = 1$}
    \STATE Set $\mathrm{update}_k \leftarrow \TRUE$ and $\klast \leftarrow k$.
  \ELSE
    \STATE Set $\mathrm{update}_k \leftarrow \FALSE$.
  \ENDIF
  \FOR{stage $h = H, H-1, \dots, 1$}
    \STATE $\hat{\wb}_{k,h} \leftarrow \bSigma_{k,h}^{-1} \sum_{i=1}^{k-1} \bar{\sigma}_{i,h}^{-2} \bphi(s_h^i, a_h^i) V_{k,h+1}(s_{h+1}^i)$.
    \STATE $\check{\wb}_{k,h} \leftarrow \bSigma_{k,h}^{-1} \sum_{i=1}^{k-1} \bar{\sigma}_{i,h}^{-2} \bphi(s_h^i, a_h^i) \check{V}_{k,h+1}(s_{h+1}^i)$.
    \STATE $\tilde{\wb}_{k,h} \leftarrow \bSigma_{k,h}^{-1} \sum_{i=1}^{k-1} \bar{\sigma}_{i,h}^{-2} \bphi(s_h^i, a_h^i) V_{k,h+1}^2(s_{h+1}^i)$.
    \IF{$\mathrm{update}_k$ is \TRUE}
      \STATE $Q_{k,h}(s,a) \leftarrow \sbr{\min\cbr{r_h(s,a) + \hat{\wb}_{k,h}^\top \bphi(s,a) + \beta \|\bphi(s,a)\|_{\bSigma_{k,h}^{-1}}, \, Q_{k-1,h}(s,a)}}_{[0,H]}$.
      \STATE $\check{Q}_{k,h}(s,a) \leftarrow \sbr{\max\cbr{r_h(s,a) + \check{\wb}_{k,h}^\top \bphi(s,a) - \bar{\beta} \|\bphi(s,a)\|_{\bSigma_{k,h}^{-1}}, \, \check{Q}_{k-1,h}(s,a)}}_{[0,H]}$.
      \STATE $V_{k,h}(s) \leftarrow \max_{a \in \cA} Q_{k,h}(s,a)$, \quad $\check{V}_{k,h}(s) \leftarrow \max_{a \in \cA} \check{Q}_{k,h}(s,a)$.
    \ELSE
      \STATE Set $Q_{k,h} \leftarrow Q_{k-1,h}$, $\check{Q}_{k,h} \leftarrow \check{Q}_{k-1,h}$, $V_{k,h} \leftarrow V_{k-1,h}$, $\check{V}_{k,h} \leftarrow \check{V}_{k-1,h}$.
    \ENDIF
  \ENDFOR
  \FOR{stage $h = 1, 2, \dots, H$}
    \STATE Take action $a_h^k \leftarrow \argmax_{a \in \cA} Q_{k,h}(s_h^k, a)$ and observe next state $s_{h+1}^k \sim \PP_h(\cdot \mid s_h^k, a_h^k)$.
    \STATE Compute empirical variance $\sigma_{k,h}$ via \eqref{eq:def-emp-var}--\eqref{eq:def-sigma-kh}.
    \STATE Set $\bar{\sigma}_{k,h} \leftarrow \max\cbr{\sigma_{k,h}, \, \gamma \|\bphi(s_h^k, a_h^k)\|_{\bSigma_{k,h}^{-1}}^{1/2}}$.
    \STATE Update $\bSigma_{k+1,h} \leftarrow \bSigma_{k,h} + \bar{\sigma}_{k,h}^{-2} \bphi(s_h^k, a_h^k)\bphi(s_h^k, a_h^k)^\top$.
  \ENDFOR
\ENDFOR
\end{algorithmic}
\end{algorithm}

\subsection{Weighted Ridge Regression and Monotonic Value Estimation}

By Proposition~\ref{prop:linear-moments}, for any optimistic value function estimate $V_{k,h+1}$, pessimistic value function estimate $\check{V}_{k,h+1}$, and squared value function $V_{k,h+1}^2$, the Bellman backups $[\PP_h V_{k,h+1}](s,a)$, $[\PP_h \check{V}_{k,h+1}](s,a)$, and $[\PP_h V_{k,h+1}^2](s,a)$ are linear in $\bphi(s,a)$ with true parameter vectors $\wb_h(V_{k,h+1})$, $\wb_h(\check{V}_{k,h+1})$, and $\wb_h(V_{k,h+1}^2)$. To estimate these vectors accurately in the presence of heteroscedastic transition noise, \algname{} solves three weighted ridge regression problems at each episode $k \in [K]$ and stage $h \in [H]$:
\begin{align*}
  \hat{\wb}_{k,h} &\leftarrow \argmin_{\wb \in \RR^d} \lambda \|\wb\|_2^2 + \sum_{i=1}^{k-1} \bar{\sigma}_{i,h}^{-2} \rbr{\wb^\top \bphi(s_h^i, a_h^i) - V_{k,h+1}(s_{h+1}^i)}^2, \\
  \check{\wb}_{k,h} &\leftarrow \argmin_{\wb \in \RR^d} \lambda \|\wb\|_2^2 + \sum_{i=1}^{k-1} \bar{\sigma}_{i,h}^{-2} \rbr{\wb^\top \bphi(s_h^i, a_h^i) - \check{V}_{k,h+1}(s_{h+1}^i)}^2, \\
  \tilde{\wb}_{k,h} &\leftarrow \argmin_{\wb \in \RR^d} \lambda \|\wb\|_2^2 + \sum_{i=1}^{k-1} \bar{\sigma}_{i,h}^{-2} \rbr{\wb^\top \bphi(s_h^i, a_h^i) - V_{k,h+1}^2(s_{h+1}^i)}^2,
\end{align*}
whose closed-form solutions are given in Lines~10--12 of Algorithm~\ref{alg:va-lsvi-ucbpp} with weighted covariance matrix $\bSigma_{k,h} = \lambda \Ib + \sum_{i=1}^{k-1} \bar{\sigma}_{i,h}^{-2} \bphi(s_h^i, a_h^i)\bphi(s_h^i, a_h^i)^\top$.

\paragraph{Monotonic value updates with rare switching.}
When the determinant-doubling criterion (Line~4) is triggered, \algname{} updates the optimistic action-value estimator $Q_{k,h}$ and pessimistic action-value estimator $\check{Q}_{k,h}$ (Lines~14--15) via
\begin{align}
  Q_{k,h}(s,a) &= \sbr{\min\cbr{r_h(s,a) + \hat{\wb}_{k,h}^\top \bphi(s,a) + \beta \|\bphi(s,a)\|_{\bSigma_{k,h}^{-1}}, \, Q_{k-1,h}(s,a)}}_{[0,H]}, \label{eq:update-Q-opt} \\
  \check{Q}_{k,h}(s,a) &= \sbr{\max\cbr{r_h(s,a) + \check{\wb}_{k,h}^\top \bphi(s,a) - \bar{\beta} \|\bphi(s,a)\|_{\bSigma_{k,h}^{-1}}, \, \check{Q}_{k-1,h}(s,a)}}_{[0,H]}. \label{eq:update-Q-pess}
\end{align}
Taking the pointwise minimum with $Q_{k-1,h}$ in \eqref{eq:update-Q-opt}, an idea originating in Algorithm~2 (\texttt{UCB-Q-values}) of \citet{azar2017minimax} for tabular MDPs (see also \citealt{zhang2020almost}) and extended to linear MDPs together with the symmetric pessimistic maximum \eqref{eq:update-Q-pess} by \citet{he2023nearly}, ensures that for every $s \in \cS$ and $h \in [H]$, $V_{k,h}(s)$ is monotonically non-increasing in $k$ while $\check{V}_{k,h}(s)$ is monotonically non-decreasing in $k$. On $\Econc$, where the sandwich holds at every stage and episode (Lemma~\ref{lem:double-induction}), this dual monotonicity guarantees that for all future episodes $j \ge k$,
\begin{align*}
  0 &\le V_{j,h+1}(s) - V_{h+1}^*(s) \le V_{k,h+1}(s) - \check{V}_{k,h+1}(s), \\
  0 &\le V_{h+1}^*(s) - \check{V}_{j,h+1}(s) \le V_{k,h+1}(s) - \check{V}_{k,h+1}(s).
\end{align*}
Consequently, a variance upper bound constructed at episode $k$ remains valid for the residual value functions $V_{j,h+1} - V_{h+1}^*$, $V_{h+1}^* - \check{V}_{j,h+1}$, and $V_{j,h+1}^2 - (V_{h+1}^*)^2$ at all future episodes $j \ge k$. Moreover, because updates only occur when $\det(\bSigma_{k,h'})$ doubles at some stage $h' \in [H]$ (Line~4), the value functions are updated at most $\cO(d H \log K)$ times, preventing the covering entropy of the function class from growing linearly with $K$.

\subsection{Variance Estimator and Confidence Radii}

At stage $h$ of episode $k$, after taking action $a_h^k$, \algname{} constructs the variance estimator $\sigma_{k,h}$ from $\hat{\wb}_{k,h}, \check{\wb}_{k,h}, \tilde{\wb}_{k,h}$ via
\begin{align}
  [\bar{\VV}_{k,h} V_{k,h+1}](s_h^k, a_h^k) &:= \sbr{\sbr{\tilde{\wb}_{k,h}^\top \bphi(s_h^k, a_h^k)}_{[0,H^2]} - \sbr{\hat{\wb}_{k,h}^\top \bphi(s_h^k, a_h^k)}_{[0,H]}^2}_{[0,H^2]}, \label{eq:def-emp-var} \\
  E_{k,h} &:= \min\cbr{\tilde{\beta}\|\bphi(s_h^k, a_h^k)\|_{\bSigma_{k,h}^{-1}}, \, H^2} + \min\cbr{2 H \bar{\beta}\|\bphi(s_h^k, a_h^k)\|_{\bSigma_{k,h}^{-1}}, \, H^2}, \label{eq:def-Ekh} \\
  D_{k,h} &:= \min\cbr{4 d^3 H^2 \rbr{(\hat{\wb}_{k,h} - \check{\wb}_{k,h})^\top \bphi(s_h^k, a_h^k) + 2\bar{\beta}\|\bphi(s_h^k, a_h^k)\|_{\bSigma_{k,h}^{-1}}}_+, \, d^3 H^3}, \label{eq:def-Dkh} \\
  \sigma_{k,h} &:= \sqrt{[\bar{\VV}_{k,h} V_{k,h+1}](s_h^k, a_h^k) + E_{k,h} + D_{k,h} + \alpha^2}. \label{eq:def-sigma-kh}
\end{align}
Here $[\bar{\VV}_{k,h} V_{k,h+1}](s_h^k, a_h^k)$ is the plug-in empirical variance of $V_{k,h+1}$, $E_{k,h}$ compensates for the estimation error between $[\bar{\VV}_{k,h} V_{k,h+1}]$ and the true variance $[\VV_h V_{k,h+1}]$, and $D_{k,h}$ bounds both the variance gap $|[\VV_h V_{k,h+1}] - [\VV_h V_{h+1}^*]|$ and $2 d^3 H$ times the residual variances $[\VV_h(V_{j,h+1} - V_{h+1}^*)]$ and $[\VV_h(V_{h+1}^* - \check{V}_{j,h+1})]$ for all $j \ge k$.

\paragraph{Variance floor and confidence radii.}
In Line~24 of Algorithm~\ref{alg:va-lsvi-ucbpp}, the regression weight $\bar{\sigma}_{k,h}$ is set as
\begin{equation}
  \bar{\sigma}_{k,h} \leftarrow \max\cbr{\sigma_{k,h}, \, \gamma \|\bphi(s_h^k, a_h^k)\|_{\bSigma_{k,h}^{-1}}^{1/2}}, \label{eq:def-sigma-bar}
\end{equation}
combining the variance estimator $\sigma_{k,h}$ (which already satisfies $\sigma_{k,h} \ge \alpha = H/\sqrt{K}$ by \eqref{eq:def-sigma-kh}) with the uncertainty-aware weight $\gamma \|\bphi(s_h^k, a_h^k)\|_{\bSigma_{k,h}^{-1}}^{1/2}$ \citep{zhou2022computationally,he2022nearly,he2023nearly}. Define the logarithmic factors
\begin{equation}
  \iota := \max\cbr{\ln\!\rbr{1 + \frac{K}{d \lambda \alpha^2}}, \, 1} = \max\cbr{\ln\!\rbr{1 + \frac{K^2}{d}}, \, 1}, \qquad \iota_0 := \ln\!\rbr{\frac{32 (d H K / \delta)^4}{\delta}}, \label{eq:def-iotas}
\end{equation}
and set the confidence radii $\beta, \bar{\beta}, \tilde{\beta}$ in Algorithm~\ref{alg:va-lsvi-ucbpp} to
\begin{equation}
  \beta = \bar{\beta} := H\sqrt{d\lambda} + 2\beta_0 + 1 = \tilde{\cO}(\sqrt{d}), \qquad \tilde{\beta} := 2 H \beta = \tilde{\cO}(H\sqrt{d}), \label{eq:def-radii}
\end{equation}
where $\beta_0 := C_0(\sqrt{d \iota^2 \iota_0} + \iota \iota_0)$ with an absolute constant $C_0 > 0$ specified in Lemma~\ref{lem:Econc}.

Comparing \eqref{eq:def-sigma-kh}--\eqref{eq:def-radii} with \lsviucbpp{} \citep{he2023nearly} (as well as \citealt{li2023variance,huang2023tackling}) highlights two closely coupled algorithmic innovations:
\begin{enumerate}
  \item \textbf{Vanishing variance floor $\alpha^2 = H^2/K$ versus constant or $K^{-1/2}$ floors.} In weighted ridge regression, a strictly positive floor $\sigma_{k,h}^2 \ge \alpha^2 > 0$ can ensure that the inverse-variance weights $\bar{\sigma}_{k,h}^{-2}$ remain well-defined when empirical variances vanish and the normalized features $\xb_{k,h} = \bar{\sigma}_{k,h}^{-1}\bphi(s_h^k, a_h^k)$ satisfy $\|\xb_{k,h}\|_2 \le \alpha^{-1}$, keeping the determinant-doubling factor $\iota$ logarithmic in $K$. However, because the cumulative exploration bonus scales with $\beta\sqrt{d H \iota \sum_{k=1}^K \sum_{h=1}^H \sigma_{k,h}^2}$ (Lemma~\ref{lem:bonus-sum}), any floor on $\sigma_{k,h}^2$ directly enters the total estimated variance $\Vest = \sum_{k=1}^K \sum_{h=1}^H \sigma_{k,h}^2$. In \lsviucbpp{} \citep{he2023nearly}, the coarse Hoeffding concentration analysis forced a constant floor $\sigma_{k,h}^2 \ge H$ (and $\bar{\sigma}_{k,h} \ge H$), making $\Vest \ge H^2 K$ and injecting an $\tilde{\cO}(d\sqrt{H^3 K})$ regret term even in deterministic MDPs; similarly, setting the floor to $\sigma_{\min}^2 \propto K^{-1/2}$ in \citet{li2023variance,huang2023tackling} still leaves an $\tilde{\cO}(K^{1/4})$ regret contribution. By decoupling the confidence radii from $\alpha^{-1}$ (which appears only inside the logarithm $\iota$), we can set $\alpha^2 = H^2/K$ in \eqref{eq:def-sigma-kh} so that the floor contributes only $\sum_{k=1}^K \sum_{h=1}^H \alpha^2 = H^3$ to $\Vest$ over all $K$ episodes.
  \item \textbf{Variance-normalized radii $\bar{\beta} = \tilde{\cO}(\sqrt{d})$ and $\tilde{\beta} = \tilde{\cO}(H\sqrt{d})$.} Because \citet{he2023nearly} analyzed $\check{\wb}_{k,h}$ and $\tilde{\wb}_{k,h}$ via a variance-oblivious Hoeffding inequality over the full function classes $\check{\cV}_{h+1}$ and $\cV_{h+1}^2$, they required large radii $\bar{\beta} = \tilde{\cO}(\sqrt{d^3 H^2})$ and $\tilde{\beta} = \tilde{\cO}(\sqrt{d^3 H^4})$ as well as $\gamma = 2 d^3 H^2$. By establishing variance-normalized Bernstein concentration for all three estimators simultaneously (Section~\ref{sec:overview}), we shave a factor of $\tilde{\cO}(d H)$ off both $\bar{\beta}$ and $\tilde{\beta}$ and tighten the uncertainty weight to $\gamma = 2 d^{1.5} H^{1.5} \iota$.
\end{enumerate}

\section{Main Results}

We now state our main theoretical guarantees for \algname{}. The complete proof of Theorem~\ref{thm:main} is provided in Appendices~\ref{app:concentration} and~\ref{app:regret}.

\begin{theorem}[Variance-Dependent Regret Bound for Linear MDPs]
\label{thm:main}
Under Definition~\ref{def:linear-mdp}, for any $\delta \in (0,1)$, with probability at least $1 - 3\delta$, the regret of \algname{} (Algorithm~\ref{alg:va-lsvi-ucbpp}) is bounded by
\begin{align}
  \Regret(K) &\le \cO\!\rbr{\beta\sqrt{d H \iota \cdot \min\cbr{\Varkstar, \Varkpi}} + d^4 H^4 \beta^2 \iota + d^4 H^4 \beta \iota^3} \nonumber \\
  &= \tilde{\cO}\!\rbr{d\sqrt{H \cdot \min\cbr{\Varkstar, \Varkpi}} + d^5 H^4}. \label{eq:main-regret-bound}
\end{align}
Moreover, the total number of value function updates (Line~4) over $K$ episodes is at most $\cO(d H \iota)$, and the total computational complexity of \algname{} is $\tilde{\cO}(d^4 H^3 |\cA| K)$.
\end{theorem}

\begin{remark}[Worst-Case Minimax Optimality and Dual Variance Adaptivity]
\label{rem:minimax-adaptivity}
Since $\Varkpi \le 4 H^2 K + 16 H^2 \ln(2/\delta)$ with probability at least $1-\delta$ (Lemma~\ref{lem:traj-var}), combining \eqref{eq:main-regret-bound} with a union bound immediately yields the worst-case regret $\tilde{\cO}(d\sqrt{H^3 K} + d^5 H^4)$ with probability at least $1 - 4\delta$, whose leading term matches the $\Omega(d\sqrt{H^3 K})$ minimax lower bound for linear MDPs \citep[Remark~23]{zhou2021nearly} (see also \citealt[Remark~5.2]{he2023nearly}) up to logarithmic factors. Moreover, whereas the transition-variance terms in prior bounds \citep{zhao2023variance,li2023variance,huang2023tackling} are defined through the optimal continuation value $V_{h+1}^*$ (with additional return-variance caps in \citealt{li2023variance,huang2023tackling}), \eqref{eq:main-regret-bound} adapts to $\min\{\Varkstar, \Varkpi\}$, where the dependence on $\Varkpi$ directly yields the first-order regret bound in Corollary~\ref{cor:first-order} via Lemma~\ref{lem:traj-var}.
\end{remark}

\begin{remark}[Deterministic and Low-Variance Regimes]
In deterministic MDPs where transitions are deterministic ($\Varkstar = \Varkpi = 0$), or more generally when $\min\{\Varkstar, \Varkpi\} = \cO(1)$, Theorem~\ref{thm:main} yields a regret bound of $\tilde{\cO}(d^5 H^4)$ free of polynomial-in-$K$ dependency, removing the $\tilde{\cO}(K^{1/4})$ polynomial overhead of \vara{} \citep{li2023variance} and \heavylsvi{} \citep{huang2023tackling}.
\end{remark}

\begin{remark}[Burn-In Overhead and Computational Complexity]
\label{rem:runtime}
Compared with the $\tilde{\cO}(d\sqrt{H^3 K} + d^7 H^8)$ bound of \lsviucbpp{} \citep{he2023nearly}, Theorem~\ref{thm:main} reduces the lower-order burn-in term by a factor of $d^2 H^4$. Computationally, because $V_{k,h+1}$ and $\check{V}_{k,h+1}$ change only at $\cO(d H \iota)$ rare-switching episodes (Line~4), between switching episodes the regression right-hand side vectors $\sum_{i=1}^{k-1}\bar{\sigma}_{i,h}^{-2}\bphi(s_h^i, a_h^i) f(s_{h+1}^i)$ ($f \in \{V_{k,h+1}, \check{V}_{k,h+1}, V_{k,h+1}^2\}$) and $\bSigma_{k,h}^{-1}$ are maintained incrementally in $\cO(d^2)$ time per step; evaluating $Q_{k,h}(s_h^k, a)$ and $\check{Q}_{k,h}(s_h^k, a)$ over the $\cO(d H \iota)$ stored functions takes $\tilde{\cO}(d^3 H |\cA|)$ time per step, and recomputing the historical regression sums at the $\cO(d H \iota)$ switching episodes takes $\tilde{\cO}(d^3 H^2 |\cA| K \cdot d H) = \tilde{\cO}(d^4 H^3 |\cA| K)$ total time over $K$ episodes \citep{he2023nearly}.
\end{remark}

Combining Theorem~\ref{thm:main} with a high-probability first-order bound on $\Varkpi$ also yields an immediate small-loss regret guarantee \citep{zanette2019tighter,wagenmaker2022first}, proved in Appendix~\ref{app:corollaries}.

\begin{corollary}[First-Order Regret]
\label{cor:first-order}
In the setting of Theorem~\ref{thm:main}, let $\Vtotalstar := \sum_{k=1}^K V_1^*(s_1^k)$ denote the cumulative optimal expected return over $K$ episodes. With probability at least $1 - 4\delta$, \algname{} satisfies
\[
  \Regret(K) \le \tilde{\cO}\!\rbr{d\sqrt{H^2 \Vtotalstar} + d^5 H^4}.
\]
\end{corollary}

\begin{remark}[Comparison with First-Order Regret Bounds]
Compared with the computationally inefficient $\tilde{\cO}(\sqrt{d^3 H^3 \Vtotalstar} + d^{3.5}H^3)$ first-order bound of \texttt{Force} \citep{wagenmaker2022first}, Corollary~\ref{cor:first-order} is computationally efficient and improves the leading term by a factor of $\sqrt{d H}$. While \citet[Corollary~5.6]{huang2023tackling} also obtained the leading first-order dependence $\tilde{\cO}(d\sqrt{H^2 K V_1^*})$ via \heavylsvi{}, their complete bound (see \citealt[Theorem~F.13]{huang2023tackling}) retains a variance- and value-independent $\tilde{\cO}(d^{3.5} H^{2.75} K^{1/4})$ polynomial additive term, whereas Corollary~\ref{cor:first-order} achieves a lower-order remainder $\tilde{\cO}(d^5 H^4)$ free of polynomial-in-$K$ dependency.
\end{remark}

\section{Overview of Key Techniques}
\label{sec:overview}

In this section, we provide a technical overview of our analysis, explaining the structural bottlenecks that prevented \lsviucbpp{} \citep{he2023nearly} from achieving variance-dependent regret and how our techniques overcome them.

\subsection{Three-Way Reference-Advantage Decomposition}

The central challenge in linear MDPs is bounding the Bellman estimation errors $|(\hat{\wb}_{k,h} - \wb_h(V_{k,h+1}))^\top \bphi(s,a)|$, $|(\check{\wb}_{k,h} - \wb_h(\check{V}_{k,h+1}))^\top \bphi(s,a)|$, and $|(\tilde{\wb}_{k,h} - \wb_h(V_{k,h+1}^2))^\top \bphi(s,a)|$. For any target function $f \in \{V_{k,h+1}, \check{V}_{k,h+1}, V_{k,h+1}^2\}$, standard ridge regression algebra reduces the estimation error to the self-normalized martingale sum
\[
  \nbr{\sum_{i=1}^{k-1} \xb_{i,h} \, \bar{\sigma}_{i,h}^{-1} \Big(f(s_{h+1}^i) - [\PP_h f](s_h^i, a_h^i)\Big)}_{\bSigma_{k,h}^{-1}}, \qquad \text{where } \xb_{i,h} := \bar{\sigma}_{i,h}^{-1} \bphi(s_h^i, a_h^i).
\]
Because $V_{k,h+1}$ and $\check{V}_{k,h+1}$ depend on all trajectories up to episode $k-1$, they are not $\cF_{i,h}$-measurable for $i < k$, necessitating a uniform convergence argument over the rare-switching function classes $\cV_{h+1}$, $\check{\cV}_{h+1}$, and $\cV_{h+1}^2$. As shown in Lemma~\ref{lem:covering}, the $\epsilon$-covering numbers of these classes satisfy $\ln|\cN_\epsilon| = \tilde{\cO}(d^3 H)$.

\paragraph{Bottleneck of coarse-to-fine concentration.}
In \citet{he2023nearly}, the authors employed a two-stage coarse-to-fine concentration strategy:
\begin{enumerate}
  \item In Stage~1 \citep[Lemma~B.1]{he2023nearly}, they applied a variance-oblivious concentration bound over the $\epsilon$-nets of $\cV_{h+1}$, $\check{\cV}_{h+1}$, and $\cV_{h+1}^2$ to obtain a ``coarse'' high-probability event $\cE$. Because this coarse bound scales with the worst-case magnitude of the normalized noise $|\bar{\sigma}_{i,h}^{-1}(f(s_{h+1}^i) - [\PP_h f](s_h^i, a_h^i))| \le \|f\|_\infty \sigma_{\min}^{-1}$ multiplied by $\sqrt{\ln|\cN_\epsilon|} = \tilde{\cO}(\sqrt{d^3 H})$, \citet{he2023nearly} used the lower bound $\bar{\sigma}_{i,h} \ge \sigma_{\min} = \sqrt{H}$ in this analysis, inflating the pessimistic and second-moment radii to $\bar{\beta} = \tilde{\cO}(\sqrt{d^3 H^2})$ and $\tilde{\beta} = \tilde{\cO}(\sqrt{d^3 H^4})$. In the subsequent variance-aware analyses of \citet{li2023variance} and \citet{huang2023tackling}, retaining this coarse concentration stage yielded initial confidence radii scaling as $\tilde{\cO}(H\sqrt{d^3 H} / \sigma_{\min})$ (and $\tilde{\cO}(H^2\sqrt{d^3 H} / \sigma_{\min})$ for the second-moment estimator); optimizing their resulting regret bounds over $\sigma_{\min}$ sets $\sigma_{\min} \propto K^{-1/4}$ to balance the $\sigma_{\min}\sqrt{K}$ bonus floor against the $\sigma_{\min}^{-1}$ radius, retaining a $\tilde{\cO}(d^{3.5} H^{2.75} K^{1/4})$ additive term in their guarantees.
  \item In Stage~2 \citep[Lemma~B.5]{he2023nearly}, conditioned on the coarse event $\cE$, they applied a reference-advantage decomposition $V_{k,h+1} = V_{h+1}^* + (V_{k,h+1} - V_{h+1}^*)$ \emph{only} to the optimistic estimator $\hat{\wb}_{k,h}$ to refine its radius to $\beta = \tilde{\cO}(\sqrt{d})$.
\end{enumerate}

\paragraph{Simultaneous three-way decomposition.}
We bypass the coarse Hoeffding stage altogether by applying reference-advantage decompositions \citep{azar2017minimax,zhang2020almost,he2023nearly} to \emph{all three} regression targets simultaneously:
\begin{align}
  V_{k,h+1} &= V_{h+1}^* + \Delta V_{k,h+1}, &&\text{where } \Delta V_{k,h+1} := V_{k,h+1} - V_{h+1}^*, \nonumber \\
  \check{V}_{k,h+1} &= V_{h+1}^* - \Delta\check{V}_{k,h+1}, &&\text{where } \Delta\check{V}_{k,h+1} := V_{h+1}^* - \check{V}_{k,h+1}, \label{eq:three-way-decomp} \\
  V_{k,h+1}^2 &= (V_{h+1}^*)^2 + \Delta(V_{k,h+1}^2), &&\text{where } \Delta(V_{k,h+1}^2) := V_{k,h+1}^2 - (V_{h+1}^*)^2. \nonumber
\end{align}
Notice how \eqref{eq:three-way-decomp} separates each target into a \emph{fixed reference component} ($V_{h+1}^*$ or $(V_{h+1}^*)^2$) that is deterministic and requires no $\epsilon$-net union bound, plus a \emph{random residual component} ($\Delta V_{k,h+1}$, $\Delta\check{V}_{k,h+1}$, or $\Delta(V_{k,h+1}^2)$) that requires an $\epsilon$-net union bound with log-covering number $\ln|\cN_\epsilon| = \tilde{\cO}(d^3 H)$. To apply the variance-aware self-normalized Bernstein inequality of \citet[Theorem~4.3]{zhou2022computationally} (Lemma~\ref{lem:bernstein-sn}) to all six components without any constant floor on $\bar{\sigma}_{i,h}$, we must control the conditional variance $\bar{\sigma}_{i,h}^{-2} [\VV_h f](s_h^i, a_h^i)$ of the normalized noise $\bar{\sigma}_{i,h}^{-1}(f(s_{h+1}^i) - [\PP_h f](s_h^i, a_h^i))$ for every historical episode $i < k$:
\begin{itemize}
  \item \textbf{Controlling the fixed squared reference $(V_{h+1}^*)^2$.} While $\bar{\sigma}_{i,h}^2 \ge [\VV_h V_{h+1}^*](s_h^i, a_h^i)$ directly normalizes the variance of $V_{h+1}^*$ to at most $1$, bounding the variance of $(V_{h+1}^*)^2$ without introducing a fourth-moment estimator requires linking $[\VV_h (V_{h+1}^*)^2]$ back to $[\VV_h V_{h+1}^*]$. We apply the conditional variance bound for squared functions (Corollary~\ref{cor:var-squared}, following \citealt[Lemma~30]{chen2021implicit}, \citealt[Lemma~34]{chen2022improved}, and \citealt[Lemma~G.8]{foster2024behavior}): since $V_{h+1}^*(s') \in [0, H]$,
  \[
    \big[\VV_h (V_{h+1}^*)^2\big](s_h^i, a_h^i) \le 4 H^2 [\VV_h V_{h+1}^*](s_h^i, a_h^i) \le 4 H^2 \bar{\sigma}_{i,h}^2.
  \]
  Thus, the normalized noise of $(V_{h+1}^*)^2$ has conditional variance bounded by $4 H^2$, yielding a tight concentration radius of $\tilde{\cO}(H\sqrt{d})$.
  \item \textbf{Controlling the three residual components across future episodes $k > i$.} For the residual components $\Delta V_{k,h+1}$, $\Delta\check{V}_{k,h+1}$, and $\Delta(V_{k,h+1}^2)$, the union bound over their $\epsilon$-nets introduces a factor of $\sqrt{\ln|\cN_\epsilon|} = \tilde{\cO}(\sqrt{d^3 H})$ in the first term of Lemma~\ref{lem:bernstein-sn}. To cancel this $\sqrt{d^3 H}$ factor, we use pointwise monotonicity ($V_{k,h+1} \le V_{i,h+1}$ and $\check{V}_{k,h+1} \ge \check{V}_{i,h+1}$ for $k > i$) and the optimism/pessimism sandwich at stage $h+1$ to bound the residual variances uniformly for all $i < k$:
  \begin{align*}
    \max\cbr{[\VV_h \Delta V_{k,h+1}](s_h^i, a_h^i), \, [\VV_h \Delta\check{V}_{k,h+1}](s_h^i, a_h^i)} &\le \frac{D_{i,h}}{2 d^3 H} \le \frac{\bar{\sigma}_{i,h}^2}{2 d^3 H}, \\
    \big[\VV_h \Delta(V_{k,h+1}^2)\big](s_h^i, a_h^i) &\le \frac{H^2 D_{i,h}}{2 d^3 H} \le \frac{H^2 \bar{\sigma}_{i,h}^2}{2 d^3 H}.
  \end{align*}
  Multiplying the normalized variance bound $1/(d^3 H)$ (resp.\ $H^2/(d^3 H)$) by the covering entropy $\ln|\cN_\epsilon| = \tilde{\cO}(d^3 H)$ inside the square root of Lemma~\ref{lem:bernstein-sn} cancels the $d^3 H$ factor. Meanwhile, the uncertainty lower bound $\bar{\sigma}_{i,h} \ge \gamma \|\bphi(s_h^i, a_h^i)\|_{\bSigma_{i,h}^{-1}}^{1/2}$ with $\gamma = 2 d^{1.5} H^{1.5}\iota$ ensures that the second-order term $2\|f\|_\infty \bar{\sigma}_{i,h}^{-2}\|\bphi(s_h^i, a_h^i)\|_{\bSigma_{i,h}^{-1}} \le 2\|f\|_\infty / \gamma^2$ in Lemma~\ref{lem:bernstein-sn} also cancels the $d^3 H$ covering entropy, independent of the variance floor $\alpha$.
\end{itemize}

\paragraph{Indicator-truncated noise and double induction.}
A subtle measurability issue arises when formalizing the above bounds: at historical episode $i < k$, the variance inequalities $[\VV_h V_{h+1}^*](s_h^i, a_h^i) \le \bar{\sigma}_{i,h}^2$ and $[\VV_h \Delta V_{k,h+1}](s_h^i, a_h^i) \le \bar{\sigma}_{i,h}^2 / (d^3 H)$ only hold when stage $h+1$ satisfies the optimism/pessimism sandwich ($\check{V}_{m,h+1} \le V_{h+1}^* \le V_{m,h+1}$) and episode $i$ satisfies the stage-$h$ Bellman concentration bounds. Conditioning on events from later stages $h+1$ would destroy the martingale property at stage $h$. To break this circularity cleanly, for any fixed deterministic function $f: \cS \to [-H^2, H^2]$ and threshold $\nu > 0$, we introduce the \emph{indicator-truncated normalized noise} (see \eqref{eq:def-truncated-noise} in Appendix~\ref{app:subsec-Econc}):
\[
  \eta_{i,h}(f, \nu) := \ind\cbr{[\VV_h f](s_h^i, a_h^i) \le \nu \bar{\sigma}_{i,h}^2} \cdot \bar{\sigma}_{i,h}^{-1}\Big(f(s_{h+1}^i) - [\PP_h f](s_h^i, a_h^i)\Big).
\]
For any fixed $f$, both $[\VV_h f](s_h^i, a_h^i)$ and $\bar{\sigma}_{i,h}$ are $\cF_{i,h}$-measurable (determined before $s_{h+1}^i$ is drawn), so the indicator prefactor is predictable and the truncated noise $\eta_{i,h}(f, \nu)$ has conditional mean zero ($\EE[\eta_{i,h}(f, \nu) \mid \cF_{i,h}] = 0$) and conditional variance at most $\nu$ ($\EE[\eta_{i,h}(f, \nu)^2 \mid \cF_{i,h}] \le \nu$). Applying Lemma~\ref{lem:bernstein-sn} to $\eta_{i,h}(f, \nu)$ for $V_{h+1}^*$, $(V_{h+1}^*)^2$, and the deterministic $\epsilon$-nets defines a single unconditional high-probability event $\Econc$ (Lemma~\ref{lem:Econc}). On any sample path in $\Econc$, we then run a purely deterministic backward induction over $h = H \to 1$ and forward induction over $k = 1 \to K$ (Lemma~\ref{lem:double-induction}): at step $(h,k)$, the induction hypotheses for stage $h+1$ and prior episodes $i < k$ imply that all indicators in $\eta_{i,h}(f, \nu)$ for $i < k$ equal $1$, establishing three-way Bellman concentration at episode $k$ without circular conditioning.

\subsection{Freedman-Type Analysis for Value Differences}

Even with tight confidence radii, a second major obstacle in \citet{he2023nearly} arises when unrolling the value differences across stages. On $\Econc$, the one-step Bellman estimation error at step $(k,h)$ is bounded by the truncated exploration bonus $B_{k,h} := 4\min\{\beta\|\bphi(s_h^k, a_h^k)\|_{\bSigma_{k,h}^{-1}}, H\}$ (and $2 B_{k,h}$ for $V_{k,h} - \check{V}_{k,h}$). Telescoping the recursive value differences from stage $h$ to $H$ bounds both the cumulative regret $\Regret(K) \le \sum_{k=1}^K (V_{k,1}(s_1^k) - V_1^{\pi^k}(s_1^k))$ (at $h=1$) and the cumulative Bellman differences $S_\pi := \sum_{k=1}^K \sum_{h=1}^H [\PP_h(V_{k,h+1} - V_{h+1}^{\pi^k})](s_h^k, a_h^k)$ and $S_{\ul} := \sum_{k=1}^K \sum_{h=1}^H [\PP_h(V_{k,h+1} - \check{V}_{k,h+1})](s_h^k, a_h^k)$ in terms of $\sum_{k,h} B_{k,h}$ and the transition martingale difference sums
\[
  \sum_{k=1}^K \sum_{h'=h}^H M_{k,h'}^\pi = \sum_{k=1}^K \sum_{h'=h}^H \rbr{[\PP_{h'}(V_{k,h'+1} - V_{h'+1}^{\pi^k})](s_{h'}^k, a_{h'}^k) - \big(V_{k,h'+1}(s_{h'+1}^k) - V_{h'+1}^{\pi^k}(s_{h'+1}^k)\big)},
\]
and analogously for $M_{k,h'}^{\ul}$. In Lemmas~C.1 and~C.2 of \citet{he2023nearly}, bounding these martingale sums via the Azuma--Hoeffding inequality first incurs an unconditional $\tilde{\cO}(\sqrt{H^3 K})$ deviation term at $h=1$ and an unconditional $\tilde{\cO}(\sqrt{H^5 K})$ deviation term in $S_\pi$ and $S_{\ul}$; because $S_\pi$ and $S_{\ul}$ are subsequently used to bound the total estimated variance $\Vest = \sum_{k,h} \sigma_{k,h}^2$ (which in turn controls $\sum_{k,h} B_{k,h} \propto \sqrt{d H \Vest}$), these variance-oblivious $\sqrt{K}$ terms propagate directly into the final regret upper bound.

To remove these $\sqrt{K}$ terms, we combine the time-uniform Freedman inequality (Lemma~\ref{lem:freedman}) with a variance self-bounding argument (Lemma~\ref{lem:freedman-value-diff}), building on techniques developed in tabular and linear RL \citep{zhang2021reinforcement,chen2021implicit,chen2022improved,foster2024behavior}. On $\Econc$, optimism and pessimism ensure that $Z_{k,h'+1}^\pi := V_{k,h'+1} - V_{h'+1}^{\pi^k}$ takes values in $[0, H]$. Because any random variable $Z \in [0, H]$ satisfies $\Var(Z) \le \EE[Z^2] \le H \EE[Z]$, the conditional variance of $M_{k,h'}^\pi$ is self-bounded by the conditional expectation of $Z_{k,h'+1}^\pi$:
\[
  \EE\big[(M_{k,h'}^\pi)^2 \mid \cF_{k,h'}\big] = [\VV_{h'} Z_{k,h'+1}^\pi](s_{h'}^k, a_{h'}^k) \le H [\PP_{h'} Z_{k,h'+1}^\pi](s_{h'}^k, a_{h'}^k).
\]
Applying Lemma~\ref{lem:freedman} with step size $\eta = 1/(2 H^2) \le 1/H$ bounds the martingale sum over $h' \ge h$ by $S_\pi / (2 H) + 2 H^2 \ln(H/\delta)$. Summing across the $H$ starting stages $h \in [H]$ yields $S_\pi / 2$, which is absorbed into the left-hand side $S_\pi$ to give $S_\pi \le 2 H \sum_{k,h} B_{k,h} + 4 H^3 \ln(H/\delta)$ (and similarly $S_{\ul} \le 4 H \sum_{k,h} B_{k,h} + 4 H^3 \ln(H/\delta)$ and $\Regret(K) \le 2\sum_{k,h} B_{k,h} + 4 H^2\ln(H/\delta)$), without any $\sqrt{K}$ concentration overhead.

\subsection{Three-Set Elliptical Bonus Summation}

Finally, to bound the exploration bonus sum $\sum_{k=1}^K \min\{\beta' \|\bphi(s_h^k, a_h^k)\|_{\bSigma_{k,h}^{-1}}, C\}$ without relying on any worst-case upper bound on $\bar{\sigma}_{k,h}^2$ (which introduced an extra $d^4 H^6$ factor in \citealt[Lemma~E.1]{he2023nearly}), we partition the episodes $k \in [K]$ at each stage $h$ into three disjoint sets according to the normalized elliptical norm $\|\xb_{k,h}\|_{\bSigma_{k,h}^{-1}}$ and the active branch in \eqref{eq:def-sigma-bar} (Lemma~\ref{lem:bonus-sum}, adapting \citealt[Lemma~4.4]{zhou2022computationally}):
\begin{enumerate}\setlength{\itemsep}{1pt}\setlength{\parskip}{0pt}\setlength{\topsep}{3pt}
  \item Episodes with large normalized uncertainty $\|\xb_{k,h}\|_{\bSigma_{k,h}^{-1}} \ge 1$, whose count is at most $2 d \iota$ by the elliptical potential lemma;
  \item Remaining episodes where the uncertainty floor dominates ($\bar{\sigma}_{k,h} = \gamma \|\bphi(s_h^k, a_h^k)\|_{\bSigma_{k,h}^{-1}}^{1/2}$), where $\|\bphi(s_h^k, a_h^k)\|_{\bSigma_{k,h}^{-1}} = \gamma^2 \|\xb_{k,h}\|_{\bSigma_{k,h}^{-1}}^2$ sums to at most $2 \gamma^2 d \iota$;
  \item Remaining episodes where $\bar{\sigma}_{k,h} = \sigma_{k,h}$, where Cauchy--Schwarz directly bounds the sum by $\beta'\sqrt{2 d \iota \sum_{k=1}^K \sigma_{k,h}^2}$.
\end{enumerate}
Combined with our tighter uncertainty weight $\gamma = 2 d^{1.5} H^{1.5} \iota$ and $\tilde{\cO}(\sqrt{d})$ confidence radii, this reduces the total estimated variance burn-in term from $\tilde{\cO}(d^{10.5} H^{16})$ in \citet{he2023nearly} to $\tilde{\cO}(d^8 H^7)$ (Lemma~\ref{lem:total-est-var}), and the final regret burn-in term from $\tilde{\cO}(d^7 H^8)$ to $\tilde{\cO}(d^5 H^4)$.

\section{Conclusion and Future Work}

In this paper, we proposed \algname{}, a computationally efficient algorithm for episodic time-inhomogeneous linear MDPs that achieves variance-dependent regret $\tilde{\cO}(d\sqrt{H \cdot \min\{\Varkstar, \Varkpi\}} + d^5 H^4)$ adapting to the smaller of the optimal and played policies' realized trajectory variances with a lower-order remainder free of polynomial-in-$K$ dependency. By introducing a simultaneous three-way reference-advantage decomposition coupled with a variance self-bounding technique, deterministic double induction, and a self-bounding Freedman-type analysis for value differences, our approach removes both the $\tilde{\cO}(K^{1/4})$ polynomial overhead of \citet{li2023variance} and \citet{huang2023tackling} and the $\tilde{\cO}(\sqrt{H^3 K})$ variance-oblivious overhead of \citet{he2023nearly}. An important direction for future work is to investigate whether horizon-free, variance-dependent regret can be achieved for time-homogeneous linear MDPs under $\sum_{h=1}^H r_h \le 1$, and whether such guarantees can be obtained in a computationally efficient manner; while \citet{zhao2023variance} established computationally efficient horizon-free variance-dependent regret for time-homogeneous linear mixture MDPs and \citet{zhang2024horizon} recently obtained a computationally inefficient worst-case horizon-free $\tilde{\cO}(d^{5.5}\sqrt{K} + d^{6.5})$ bound for time-homogeneous linear MDPs (see Appendix~\ref{app:homog-comparison}), designing a variance-dependent or computationally efficient horizon-free algorithm for time-homogeneous linear MDPs remains open.

\section*{Acknowledgement}

The proof ideas are developed through an interactive brainstorm between Gemini 4 Argon and the authors. Gemini 4 Argon and Gemini 4 Carbon are employed to further streamline the canonical pillars underlying the proof roadmap. The authors simplify the algorithms and proof techniques, re-organize and verify the proofs, and prepare the draft with further assistance from Gemini 4 Carbon and GPT-6 Astra. The authors take full responsibility for its contents.

\newpage
\appendix

\section{Concentration Analysis and Master Double Induction}
\label{app:concentration}

In this appendix, we establish the rare-switching update bound, the $\epsilon$-covering numbers of the shifted residual function classes, the unconditional high-probability concentration event $\Econc$, and the master deterministic double induction lemma (Lemma~\ref{lem:double-induction}).

\subsection{Function Classes and Covering Numbers}

We begin by bounding the number of episodes in which the determinant-doubling condition (Line~4 of Algorithm~\ref{alg:va-lsvi-ucbpp}) is triggered.

\begin{lemma}[Number of Value Function Updates]
\label{lem:switching-count}
In Algorithm~\ref{alg:va-lsvi-ucbpp}, the total number of episodes $k \in [K]$ in which $\mathrm{update}_k = \textup{\textbf{true}}$ (Line~4) is bounded by
\[
  1 + \frac{d H}{\ln 2}\ln\!\rbr{1 + \frac{K}{d \lambda \alpha^2}} \le 1 + \frac{d H \iota}{\ln 2} =: \lmax.
\]
\end{lemma}

\begin{proof}
Let $1 = k_1 < k_2 < \dots < k_m \le K$ denote the episodes in which Line~4 is triggered. By the update rule in Line~4, for each $j \in \{2, \dots, m\}$, there exists at least one stage $h' \in [H]$ such that $\det(\bSigma_{k_j, h'}) \ge 2 \det(\bSigma_{k_{j-1}, h'})$. Since $\bSigma_{k,h}$ is non-decreasing in $k$ in the Loewner order for every stage $h \in [H]$, we have $\prod_{h=1}^H \det(\bSigma_{k_j, h}) \ge 2 \prod_{h=1}^H \det(\bSigma_{k_{j-1}, h})$ for all $j \in \{2, \dots, m\}$. Unrolling this inequality over $j = m, m-1, \dots, 2$ gives
\begin{equation}
  \prod_{h=1}^H \det(\bSigma_{K,h}) \ge \prod_{h=1}^H \det(\bSigma_{k_m, h}) \ge 2^{m-1} \prod_{h=1}^H \det(\bSigma_{1,h}) = 2^{m-1} \lambda^{d H}, \label{eq:det-lower}
\end{equation}
where the first inequality holds because $\bSigma_{K,h} \succeq \bSigma_{k_m,h}$ and the equality uses $\bSigma_{1,h} = \lambda \Ib$. On the other hand, for every stage $h \in [H]$, we have
\begin{equation}
  \det(\bSigma_{K,h}) \le \rbr{\frac{\tr(\bSigma_{K,h})}{d}}^d \le \rbr{\frac{d\lambda + \sum_{k=1}^{K-1} \bar{\sigma}_{k,h}^{-2}\|\bphi(s_h^k, a_h^k)\|_2^2}{d}}^d \le \lambda^d \rbr{1 + \frac{K}{d \lambda \alpha^2}}^d \le \lambda^d e^{d\iota}, \label{eq:det-upper}
\end{equation}
where the first inequality follows from the arithmetic--geometric mean inequality, the third inequality uses $\|\bphi(s_h^k, a_h^k)\|_2 \le 1$ and $\bar{\sigma}_{k,h} \ge \sigma_{k,h} \ge \alpha$ from \eqref{eq:def-sigma-kh}--\eqref{eq:def-sigma-bar}, and the last inequality uses \eqref{eq:def-iotas}. Taking the product of \eqref{eq:det-upper} over $h \in [H]$ and combining with \eqref{eq:det-lower} gives $2^{m-1}\lambda^{d H} \le \lambda^{d H} e^{d H \iota}$, which implies $m \le 1 + d H \iota / \ln 2 = \lmax$.
\end{proof}

Next, we establish unconditional Euclidean norm bounds for the regression weight vectors $\hat{\wb}_{k,h}$, $\check{\wb}_{k,h}$, and $\tilde{\wb}_{k,h}$.

\begin{lemma}[Norm Bounds for Ridge Regression Weights]
\label{lem:weight-norms}
For all $k \in [K]$ and $h \in [H]$, the regression weight vectors in Lines~10--12 of Algorithm~\ref{alg:va-lsvi-ucbpp} unconditionally satisfy
\[
  \|\hat{\wb}_{k,h}\|_2 \le L, \qquad \|\check{\wb}_{k,h}\|_2 \le L, \qquad \|\tilde{\wb}_{k,h}\|_2 \le H L, \qquad \text{where } L := \frac{H\sqrt{d K}}{\alpha\sqrt{\lambda}}.
\]
\end{lemma}

\begin{proof}
Consider any vector of the form $\wb = \bSigma_{k,h}^{-1}\sum_{i=1}^{k-1} \bar{\sigma}_{i,h}^{-2} \bphi(s_h^i, a_h^i) y_i$ where $y_i \in [0, B]$ for all $i < k$ (with $B = H$ for $\hat{\wb}_{k,h}, \check{\wb}_{k,h}$ and $B = H^2$ for $\tilde{\wb}_{k,h}$). Let $\bphi_i := \bphi(s_h^i, a_h^i)$. Then we have
\begin{align*}
  \|\wb\|_2^2 &= \nbr{\sum_{i=1}^{k-1} \bar{\sigma}_{i,h}^{-2} y_i \bSigma_{k,h}^{-1} \bphi_i}_2^2 \le (k-1) \sum_{i=1}^{k-1} \bar{\sigma}_{i,h}^{-4} y_i^2 \bphi_i^\top \bSigma_{k,h}^{-2} \bphi_i \\
  &\le \frac{K B^2}{\lambda \alpha^2} \sum_{i=1}^{k-1} \bar{\sigma}_{i,h}^{-2} \bphi_i^\top \bSigma_{k,h}^{-1} \bphi_i = \frac{K B^2}{\lambda \alpha^2} \tr\!\rbr{\bSigma_{k,h}^{-1}(\bSigma_{k,h} - \lambda \Ib)} \le \frac{d K B^2}{\lambda \alpha^2},
\end{align*}
where the first inequality follows from the Cauchy--Schwarz inequality, the second inequality holds since $y_i \in [0,B]$, $\bar{\sigma}_{i,h} \ge \alpha$, and $\bSigma_{k,h}^{-1} \preceq \lambda^{-1}\Ib$, and the last inequality uses $\bSigma_{k,h} - \lambda\Ib \preceq \bSigma_{k,h}$ in $\RR^{d\times d}$. Taking the square root with $B = H$ and $B = H^2$ completes the proof.
\end{proof}

By Lemmas~\ref{lem:switching-count} and~\ref{lem:weight-norms} and the truncation $[\cdot]_{[0,H]}$ in Lines~14--15 of Algorithm~\ref{alg:va-lsvi-ucbpp}, for every $k \in [K]$ and $h \in [H]$, the optimistic and pessimistic value functions $V_{k,h}$ and $\check{V}_{k,h}$ unconditionally take values in $[0,H]$ and belong to the function classes
\begin{align}
  \cV_h &:= \cbr{\max_{a\in\cA}\min_{1 \le i \le \lmax}\sbr{r_h(\cdot, a) + \wb_i^\top \bphi(\cdot, a) + \beta\|\bphi(\cdot, a)\|_{\bGamma_i}}_{[0,H]} : \|\wb_i\|_2 \le L, \, \bzero \preceq \bGamma_i \preceq \frac{1}{\lambda}\Ib}, \label{eq:def-Vh} \\
  \check{\cV}_h &:= \cbr{\max_{a\in\cA}\max_{1 \le i \le \lmax}\sbr{r_h(\cdot, a) + \wb_i^\top \bphi(\cdot, a) - \bar{\beta}\|\bphi(\cdot, a)\|_{\bGamma_i}}_{[0,H]} : \|\wb_i\|_2 \le L, \, \bzero \preceq \bGamma_i \preceq \frac{1}{\lambda}\Ib}. \label{eq:def-Vh-check}
\end{align}
To analyze the three-way reference-advantage decomposition in \eqref{eq:three-way-decomp}, we define the three shifted residual function classes at stage $h+1$ from \eqref{eq:def-Vh}--\eqref{eq:def-Vh-check} as
\begin{equation}
\begin{aligned}
  \Delta\cV_{h+1} &:= \cbr{V - V_{h+1}^* : V \in \cV_{h+1}}, \qquad \Delta\check{\cV}_{h+1} := \cbr{V_{h+1}^* - \check{V} : \check{V} \in \check{\cV}_{h+1}}, \\
  \Delta(\cV_{h+1}^2) &:= \cbr{V^2 - (V_{h+1}^*)^2 : V \in \cV_{h+1}}.
\end{aligned}
\label{eq:def-residual-classes}
\end{equation}
Because every $V \in \cV_{h+1}$ and $\check{V} \in \check{\cV}_{h+1}$ takes values in $[0,H]$, every $f \in \Delta\cV_{h+1} \cup \Delta\check{\cV}_{h+1}$ satisfies $\|f\|_\infty \le H$, and every $f \in \Delta(\cV_{h+1}^2)$ satisfies $\|f\|_\infty \le H^2$.

\begin{lemma}[$\epsilon$-Covering Numbers of Residual Function Classes]
\label{lem:covering}
Let $\dist(f_1, f_2) := \|f_1 - f_2\|_\infty$. Set the covering radii
\begin{align}
  \epsilon_1 &:= \frac{\alpha^2\sqrt{\lambda}}{16 d^3 H^2 K} = \frac{1}{16 d^3 H K^2}, \label{eq:def-eps1} \\
  \epsilon_2 &:= 2 H \epsilon_1 = \frac{\alpha^2\sqrt{\lambda}}{8 d^3 H K} = \frac{1}{8 d^3 K^2}. \label{eq:def-eps2}
\end{align}
Then for every stage $h \in [H]$, the residual classes in \eqref{eq:def-residual-classes} admit deterministic nets $\cN_{\epsilon_1}(\Delta\cV_{h+1}) \subset \Delta\cV_{h+1}$, $\cN_{\epsilon_1}(\Delta\check{\cV}_{h+1}) \subset \Delta\check{\cV}_{h+1}$, and $\cN_{\epsilon_2}(\Delta(\cV_{h+1}^2)) \subset \Delta(\cV_{h+1}^2)$ (with radii $\epsilon_1, \epsilon_1, \epsilon_2$) satisfying
\[
  \max\cbr{\ln\abr{\cN_{\epsilon_1}(\Delta\cV_{h+1})}, \, \ln\abr{\cN_{\epsilon_1}(\Delta\check{\cV}_{h+1})}, \, \ln\abr{\cN_{\epsilon_2}(\Delta(\cV_{h+1}^2))}} \le C_{\cN} d^3 H \iota \iota_0,
\]
where $C_{\cN} > 0$ is an absolute constant.
\end{lemma}

\begin{proof}
Since $V_{h+1}^*$ and $(V_{h+1}^*)^2$ are fixed deterministic functions, shifting by $V_{h+1}^*$ or $(V_{h+1}^*)^2$ is an isometry under $\dist(\cdot, \cdot)$. Moreover, for any $V_1, V_2 \in \cV_{h+1}$, since $V_1(s), V_2(s) \in [0,H]$, we have $|V_1^2(s) - V_2^2(s)| = |V_1(s) + V_2(s)| \cdot |V_1(s) - V_2(s)| \le 2 H \dist(V_1, V_2)$. Hence any $\epsilon_1$-net of $\cV_{h+1}$ directly induces an $\epsilon_2$-net of $\Delta(\cV_{h+1}^2)$ with $\epsilon_2 = 2 H \epsilon_1$ of identical cardinality.

To bound the $\epsilon_1$-covering number of $\cV_{h+1}$ (and identically of $\check{\cV}_{h+1}$ since $\bar{\beta} = \beta$), observe that $\max_{a\in\cA}$, $\min_{1\le i\le \lmax}$ (or $\max_{1\le i\le \lmax}$), and $[\cdot]_{[0,H]}$ are all $1$-Lipschitz under $\|\cdot\|_\infty$. Thus, for any $V_1, V_2 \in \cV_{h+1}$ parameterized by $\{(\wb_{1,i}, \bGamma_{1,i})\}_{i=1}^{\lmax}$ and $\{(\wb_{2,i}, \bGamma_{2,i})\}_{i=1}^{\lmax}$, we have
\begin{align}
  \dist(V_1, V_2)
  &\le \max_{1 \le i \le \lmax} \sup_{s\in\cS, a\in\cA} \abr{(\wb_{1,i} - \wb_{2,i})^\top \bphi(s,a) + \beta\rbr{\|\bphi(s,a)\|_{\bGamma_{1,i}} - \|\bphi(s,a)\|_{\bGamma_{2,i}}}} \nonumber \\
  &\le \max_{1 \le i \le \lmax} \|\wb_{1,i} - \wb_{2,i}\|_2 + \beta \max_{1 \le i \le \lmax} \sqrt{\|\bGamma_{1,i} - \bGamma_{2,i}\|_F}, \label{eq:covering-lip}
\end{align}
where the second inequality follows from $\|\bphi(s,a)\|_2 \le 1$, $|\sqrt{u} - \sqrt{v}| \le \sqrt{|u-v|}$ for $u, v \ge 0$, and the Cauchy--Schwarz inequality. Notice that $\mathbb{B}_2^d(L) := \{\wb \in \RR^d : \|\wb\|_2 \le L\}$ is a Euclidean ball of radius $L$, and $\mathbb{S}_+^d(\lambda^{-1}) := \{\bGamma \in \RR^{d\times d} : \bzero \preceq \bGamma \preceq \lambda^{-1}\Ib\}$ is contained in the Frobenius ball of radius $\sqrt{d}/\lambda$ in $\RR^{d\times d}$. Since the $\epsilon$-covering number of any subset of a normed ball of radius $R$ (by elements within the subset) is bounded by the $(\epsilon/2)$-covering number of the ambient ball, Lemma~\ref{lem:euclidean-covering} implies that $\mathbb{B}_2^d(L)$ admits an $(\epsilon_1/2)$-net of size at most $(1 + 4L/\epsilon_1)^d$ and $\mathbb{S}_+^d(\lambda^{-1})$ admits an $(\epsilon_1^2/(4\beta^2))$-net under $\|\cdot\|_F$ of size at most $(1 + 16\sqrt{d}\beta^2/(\lambda\epsilon_1^2))^{d^2}$. Taking the product over $1 \le i \le \lmax$, it follows from \eqref{eq:covering-lip} that
\[
  \ln\abr{\cN_{\epsilon_1}(\Delta\cV_{h+1})} \le d \lmax \ln\!\rbr{1 + \frac{4 L}{\epsilon_1}} + d^2 \lmax \ln\!\rbr{1 + \frac{16\sqrt{d}\beta^2}{\lambda\epsilon_1^2}} \le C_{\cN} d^3 H \iota \iota_0,
\]
where the second inequality uses $\lmax = \cO(d H \iota)$ (Lemma~\ref{lem:switching-count}) and $\ln(1 + 4L/\epsilon_1) + \ln(1 + 16\sqrt{d}\beta^2/(\lambda\epsilon_1^2)) = \cO(\iota_0)$.
\end{proof}

\subsection{High-Probability Concentration Event}
\label{app:subsec-Econc}

Recall that $\cF_{i,h}$ denotes the $\sigma$-algebra generated by all random variables up to $(s_h^i, a_h^i)$ prior to observing $s_{h+1}^i$. By construction of Algorithm~\ref{alg:va-lsvi-ucbpp}, $\bar{\sigma}_{i,h}$, $\xb_{i,h} := \bar{\sigma}_{i,h}^{-1}\bphi(s_h^i, a_h^i)$, and $\bSigma_{i,h}$ are all $\cF_{i,h}$-measurable. For any fixed deterministic function $f: \cS \to [-H^2, H^2]$ and any deterministic variance threshold $\nu > 0$, we define the indicator-truncated normalized noise:
\begin{equation}
  \eta_{i,h}(f, \nu) := \ind\cbr{[\VV_h f](s_h^i, a_h^i) \le \nu \bar{\sigma}_{i,h}^2} \cdot \bar{\sigma}_{i,h}^{-1}\Big(f(s_{h+1}^i) - [\PP_h f](s_h^i, a_h^i)\Big). \label{eq:def-truncated-noise}
\end{equation}
Because the indicator $\ind\{[\VV_h f](s_h^i, a_h^i) \le \nu \bar{\sigma}_{i,h}^2\}$ is $\cF_{i,h}$-measurable and $\eta_{i,h}(f, \nu)$ is observed once $s_{h+1}^i$ is drawn ($\cF_{i+1,h}$-measurable), $\eta_{i,h}(f, \nu)$ has conditional mean zero ($\EE[\eta_{i,h}(f, \nu) \mid \cF_{i,h}] = 0$) and conditional variance at most $\nu$ ($\EE[\eta_{i,h}(f, \nu)^2 \mid \cF_{i,h}] \le \nu$). This allows us to apply the variance-aware self-normalized concentration inequality (Lemma~\ref{lem:bernstein-sn}) with filtration $\{\cF_{i,h}\}_{i\ge 1}$ to establish the following high-probability event $\Econc$.

\begin{lemma}[Unconditional High-Probability Concentration Event $\Econc$]
\label{lem:Econc}
Define $\Econc$ as the event that for all stages $h \in [H]$ and all episodes $k \in [K]$, the following five inequalities hold simultaneously:
\begin{align}
  \nbr{\sum_{i=1}^{k-1} \xb_{i,h} \eta_{i,h}\big(V_{h+1}^*, 1\big)}_{\bSigma_{k,h}^{-1}} &\le \beta_0, \label{eq:Econc-Vstar} \\
  \nbr{\sum_{i=1}^{k-1} \xb_{i,h} \eta_{i,h}\big((V_{h+1}^*)^2, 4 H^2\big)}_{\bSigma_{k,h}^{-1}} &\le 2 H \beta_0, \label{eq:Econc-Vstar-sq} \\
  \sup_{\tilde{V} \in \cN_{\epsilon_1}(\Delta\cV_{h+1})} \nbr{\sum_{i=1}^{k-1} \xb_{i,h} \eta_{i,h}\!\rbr{\tilde{V}, \frac{1}{d^3 H}}}_{\bSigma_{k,h}^{-1}} &\le \beta_0, \label{eq:Econc-net-opt} \\
  \sup_{\tilde{U} \in \cN_{\epsilon_1}(\Delta\check{\cV}_{h+1})} \nbr{\sum_{i=1}^{k-1} \xb_{i,h} \eta_{i,h}\!\rbr{\tilde{U}, \frac{1}{d^3 H}}}_{\bSigma_{k,h}^{-1}} &\le \beta_0, \label{eq:Econc-net-pess} \\
  \sup_{\tilde{W} \in \cN_{\epsilon_2}(\Delta(\cV_{h+1}^2))} \nbr{\sum_{i=1}^{k-1} \xb_{i,h} \eta_{i,h}\!\rbr{\tilde{W}, \frac{H^2}{d^3 H}}}_{\bSigma_{k,h}^{-1}} &\le H \beta_0, \label{eq:Econc-net-sq}
\end{align}
where $\beta_0 := C_0(\sqrt{d \iota^2 \iota_0} + \iota \iota_0)$ with the explicit absolute constant $C_0 := 12\sqrt{C_{\cN} + 2} + 60(C_{\cN} + 2)$. Then $\PP(\Econc) \ge 1 - \delta$.
\end{lemma}

\begin{proof}
Fix any stage $h \in [H]$, any deterministic function $f: \cS \to [-H^2, H^2]$, and any threshold $\nu > 0$. Since $\ind\{[\VV_h f](s_h^i, a_h^i) \le \nu \bar{\sigma}_{i,h}^2\}$ is $\cF_{i,h}$-measurable, the sequence $\{\xb_{i,h}, \eta_{i,h}(f, \nu)\}_{i=1}^K$ satisfies the conditions of Lemma~\ref{lem:bernstein-sn} with $\EE[\eta_{i,h}(f,\nu) \mid \cF_{i,h}] = 0$ and
\begin{align*}
  \EE\big[\eta_{i,h}(f,\nu)^2 \mid \cF_{i,h}\big] &= \ind\cbr{[\VV_h f](s_h^i, a_h^i) \le \nu \bar{\sigma}_{i,h}^2} \bar{\sigma}_{i,h}^{-2} [\VV_h f](s_h^i, a_h^i) \le \nu, \\
  \|\xb_{i,h}\|_2 &\le \alpha^{-1}, \qquad |\eta_{i,h}(f, \nu)| \le 2 \|f\|_\infty \alpha^{-1} \le 2 H^2 \alpha^{-1} =: R.
\end{align*}
Furthermore, for all $k \in [K]$, the second-order term in Lemma~\ref{lem:bernstein-sn} satisfies
\begin{equation}
  \max_{1 \le i \le k-1} |\eta_{i,h}(f, \nu)| \min\cbr{1, \|\xb_{i,h}\|_{\bSigma_{i,h}^{-1}}} \le \max_{1 \le i \le k-1} 2 \|f\|_\infty \bar{\sigma}_{i,h}^{-2} \|\bphi(s_h^i, a_h^i)\|_{\bSigma_{i,h}^{-1}} \le \frac{2 \|f\|_\infty}{\gamma^2}, \label{eq:second-order-cancel}
\end{equation}
where the second inequality uses $\bar{\sigma}_{i,h} \ge \gamma \|\bphi(s_h^i, a_h^i)\|_{\bSigma_{i,h}^{-1}}^{1/2}$ from \eqref{eq:def-sigma-bar}. We set the discretization parameter in Lemma~\ref{lem:bernstein-sn} to $\epsilon_{\mathrm{c}} := 1/(d^3 H^2)$ and note that $\ln(1 + K \alpha^{-2}/(d\lambda)) \le \iota$ and $\iota \ge 1$ by \eqref{eq:def-iotas}.

To prove \eqref{eq:Econc-Vstar} and \eqref{eq:Econc-Vstar-sq}, we apply Lemma~\ref{lem:bernstein-sn} to $f = V_{h+1}^*$ (with $\nu = 1$, $\|f\|_\infty \le H$) and to $f = (V_{h+1}^*)^2$ (with $\nu = 4 H^2$, $\|f\|_\infty \le H^2$), each with failure probability $\delta' = \delta/(5 H)$. Then $\iota_{\mathrm{c}}(\delta') = \ln(32(\ln(R/\epsilon_{\mathrm{c}}) + 1)K^2/\delta') \le 2\iota_0$. For $f = V_{h+1}^*$, with probability at least $1 - \delta/(5H)$, simultaneously for all $k \in [K]$:
\[
  \nbr{\sum_{i=1}^{k-1} \xb_{i,h} \eta_{i,h}\big(V_{h+1}^*, 1\big)}_{\bSigma_{k,h}^{-1}} \le 12\sqrt{2 d \iota \iota_0} + 48\iota_0 \cdot \frac{2 H}{\gamma^2} + 12\iota_0 \epsilon_{\mathrm{c}} \le \big(12\sqrt{2} + 60\big)\big(\sqrt{d\iota^2\iota_0} + \iota\iota_0\big) \le \beta_0,
\]
where the first inequality follows from Lemma~\ref{lem:bernstein-sn} and \eqref{eq:second-order-cancel}, and the second uses $2 H / \gamma^2 \le 1$, $\epsilon_{\mathrm{c}} \le 1$, and $\iota \ge 1$. Similarly, for $f = (V_{h+1}^*)^2$ with $\nu = 4 H^2$ and $\|f\|_\infty \le H^2$, with probability at least $1 - \delta/(5H)$, simultaneously for all $k \in [K]$:
\[
  \nbr{\sum_{i=1}^{k-1} \xb_{i,h} \eta_{i,h}\big((V_{h+1}^*)^2, 4 H^2\big)}_{\bSigma_{k,h}^{-1}} \le 24 H\sqrt{2 d \iota \iota_0} + 48\iota_0 \cdot \frac{2 H^2}{\gamma^2} + 12\iota_0 \epsilon_{\mathrm{c}} \le 2 H \beta_0,
\]
where the first inequality applies Lemma~\ref{lem:bernstein-sn} and \eqref{eq:second-order-cancel}, and the second uses $2 H^2 / \gamma^2 \le H$ and $\epsilon_{\mathrm{c}} \le 1 \le H$.

To prove \eqref{eq:Econc-net-opt}, \eqref{eq:Econc-net-pess}, and \eqref{eq:Econc-net-sq}, we apply Lemma~\ref{lem:bernstein-sn} to each fixed function in the corresponding deterministic $\epsilon$-net $\cN$ with failure probability $\delta' = \delta / (5 H |\cN|)$ and take a union bound over $\cN$. By Lemma~\ref{lem:covering}, we have $\iota_{\mathrm{c}}(\delta') \le \ln|\cN| + 2\iota_0 \le (C_{\cN} + 2) d^3 H \iota \iota_0$. For $\cN_{\epsilon_1}(\Delta\cV_{h+1})$ (and identically for $\cN_{\epsilon_1}(\Delta\check{\cV}_{h+1})$), with probability at least $1 - \delta/(5H)$, simultaneously for all $k \in [K]$:
\begin{align*}
  &\sup_{\tilde{V} \in \cN_{\epsilon_1}(\Delta\cV_{h+1})} \nbr{\sum_{i=1}^{k-1} \xb_{i,h} \eta_{i,h}\!\rbr{\tilde{V}, \frac{1}{d^3 H}}}_{\bSigma_{k,h}^{-1}} \\
  &\quad \le 12\sqrt{\frac{1}{d^3 H} \cdot d\iota \cdot (C_{\cN} + 2) d^3 H \iota \iota_0} + 24(C_{\cN} + 2) d^3 H \iota \iota_0 \cdot \frac{2 H}{\gamma^2} + 6(C_{\cN} + 2) d^3 H \iota \iota_0 \cdot \epsilon_{\mathrm{c}} \\
  &\quad = 12\sqrt{C_{\cN} + 2}\sqrt{d \iota^2 \iota_0} + 12(C_{\cN} + 2)\frac{\iota_0}{H \iota} + 6(C_{\cN} + 2)\frac{\iota \iota_0}{H} \\
  &\quad \le C_0\big(\sqrt{d\iota^2\iota_0} + \iota\iota_0\big) = \beta_0,
\end{align*}
where the first inequality applies Lemma~\ref{lem:bernstein-sn} and \eqref{eq:second-order-cancel} with $\nu = 1/(d^3 H)$ and $\|\tilde{V}\|_\infty \le H$, the equality substitutes $\gamma^2 = 4 d^3 H^3 \iota^2$ and $\epsilon_{\mathrm{c}} = 1/(d^3 H^2)$, and the last inequality uses $H \ge 1$ and $\iota \ge 1$. Finally, for $\cN_{\epsilon_2}(\Delta(\cV_{h+1}^2))$ in \eqref{eq:Econc-net-sq}, replacing $(\nu, \|\tilde{V}\|_\infty) = (1/(d^3 H), H)$ by $(H^2/(d^3 H), H^2)$ multiplies each of the three terms above by at most $H$, yielding an upper bound of $H \beta_0$ with probability at least $1 - \delta/(5H)$. Taking a union bound over all $5 H$ events completes the proof.
\end{proof}

\subsection{Master Double Induction}

We now prove that on the single event $\Econc$, all three regression estimators enjoy Bernstein concentration, the variance estimator $\sigma_{k,h}^2$ upper-bounds $[\VV_h V_{h+1}^*](s_h^k, a_h^k)$ up to known error terms, the residual variances remain uniformly bounded for all future episodes $j \ge k$, and the value function estimates sandwich $Q_h^*, V_h^*$.

\begin{lemma}[Master Double Induction Lemma]
\label{lem:double-induction}
On the event $\Econc$, for every stage $h \in [H]$ and every episode $k \in [K]$, the following four properties hold simultaneously:
\begin{enumerate}
  \item[$\textup{\textbf{(P1)}}_{k,h}$] \textbf{Three-way Bellman concentration.} For all $(s,a) \in \cS \times \cA$,
  \begin{align}
    \abr{\hat{\wb}_{k,h}^\top \bphi(s,a) - [\PP_h V_{k,h+1}](s,a)} &\le \beta \|\bphi(s,a)\|_{\bSigma_{k,h}^{-1}}, \label{eq:P1-opt} \\
    \abr{\check{\wb}_{k,h}^\top \bphi(s,a) - [\PP_h \check{V}_{k,h+1}](s,a)} &\le \bar{\beta} \|\bphi(s,a)\|_{\bSigma_{k,h}^{-1}}, \label{eq:P1-pess} \\
    \abr{\tilde{\wb}_{k,h}^\top \bphi(s,a) - [\PP_h V_{k,h+1}^2](s,a)} &\le \tilde{\beta} \|\bphi(s,a)\|_{\bSigma_{k,h}^{-1}}. \label{eq:P1-sq}
  \end{align}
  \item[$\textup{\textbf{(P2)}}_{k,h}$] \textbf{Variance estimation accuracy.}
  \begin{align}
    \abr{[\bar{\VV}_{k,h} V_{k,h+1}](s_h^k, a_h^k) - [\VV_h V_{k,h+1}](s_h^k, a_h^k)} &\le E_{k,h}, \label{eq:P2-Vk} \\
    \abr{[\bar{\VV}_{k,h} V_{k,h+1}](s_h^k, a_h^k) - [\VV_h V_{h+1}^*](s_h^k, a_h^k)} &\le E_{k,h} + D_{k,h}, \label{eq:P2-Vstar}
  \end{align}
  and consequently $\bar{\sigma}_{k,h}^2 \ge \sigma_{k,h}^2 \ge [\VV_h V_{h+1}^*](s_h^k, a_h^k) + \alpha^2$.
  \item[$\textup{\textbf{(P3)}}_{k,h}$] \textbf{Uniform residual variance bounds.} For every $j \in \{k, k+1, \dots, K\}$,
  \begin{align}
    \big[\VV_h(V_{j,h+1} - V_{h+1}^*)\big](s_h^k, a_h^k) &\le \frac{D_{k,h}}{2 d^3 H}, \label{eq:P3-opt} \\
    \big[\VV_h(V_{h+1}^* - \check{V}_{j,h+1})\big](s_h^k, a_h^k) &\le \frac{D_{k,h}}{2 d^3 H}, \label{eq:P3-pess} \\
    \big[\VV_h\big(V_{j,h+1}^2 - (V_{h+1}^*)^2\big)\big](s_h^k, a_h^k) &\le \frac{H^2 D_{k,h}}{2 d^3 H}. \label{eq:P3-sq}
  \end{align}
  \item[$\textup{\textbf{(P4)}}_{k,h}$] \textbf{Optimism and pessimism sandwich.} For all $(s,a) \in \cS \times \cA$,
  \[
    \check{Q}_{k,h}(s,a) \le Q_h^*(s,a) \le Q_{k,h}(s,a), \qquad \check{V}_{k,h}(s) \le V_h^*(s) \le V_{k,h}(s).
  \]
\end{enumerate}
\end{lemma}

\begin{proof}
Fix any sample path in $\Econc$. We proceed by outer backward induction on $h = H, H-1, \dots, 1$ and inner forward induction on $k = 1, 2, \dots, K$.

\paragraph{Outer induction hypothesis.}
Fix stage $h \in [H]$ and assume that at stage $h+1$, property~$\textup{\textbf{(P4)}}_{m,h+1}$ holds for all episodes $m \in [K]$:
\begin{equation}
  0 \le \check{V}_{m,h+1}(s) \le V_{h+1}^*(s) \le V_{m,h+1}(s) \le H \qquad \forall s \in \cS, \, m \in [K]. \label{eq:outer-ind-hyp}
\end{equation}
For the base case $h = H$, $V_{m,H+1}(s) = V_{H+1}^*(s) = \check{V}_{m,H+1}(s) = 0$, so \eqref{eq:outer-ind-hyp} holds trivially.

\paragraph{Inner induction on $k = 1, \dots, K$ for $\textup{(P1)}_{k,h}$--$\textup{(P3)}_{k,h}$.}
Fix $k \in [K]$ and assume that for all $i \in \{1, \dots, k-1\}$ (vacuously true when $k = 1$), properties~$\textup{\textbf{(P1)}}_{i,h}$, $\textup{\textbf{(P2)}}_{i,h}$, and~$\textup{\textbf{(P3)}}_{i,h}$ hold.

\paragraph{Step 1: Proof of $\textup{(P1)}_{k,h}$.}
Consider any target function $f \in \{V_{k,h+1}, \check{V}_{k,h+1}, V_{k,h+1}^2\}$ with true Bellman backup parameter $\wb_h(f)$ (Proposition~\ref{prop:linear-moments}) and weighted ridge estimator $\hat{\wb}(f) := \bSigma_{k,h}^{-1}\sum_{i=1}^{k-1}\bar{\sigma}_{i,h}^{-2}\bphi(s_h^i, a_h^i) f(s_{h+1}^i)$. For all $(s,a) \in \cS \times \cA$, we have
\begin{align}
  \abr{\hat{\wb}(f)^\top \bphi(s,a) - [\PP_h f](s,a)}
  &= \abr{\inner{\hat{\wb}(f) - \wb_h(f)}{\bphi(s,a)}} \nonumber \\
  &\le \|\hat{\wb}(f) - \wb_h(f)\|_{\bSigma_{k,h}} \|\bphi(s,a)\|_{\bSigma_{k,h}^{-1}}, \label{eq:ridge-cs}
\end{align}
where the equality uses $[\PP_h f](s,a) = \wb_h(f)^\top \bphi(s,a)$ and the inequality follows from the Cauchy--Schwarz inequality. Expanding $\wb_h(f) = \bSigma_{k,h}^{-1}(\lambda\Ib + \sum_{i=1}^{k-1}\bar{\sigma}_{i,h}^{-2}\bphi(s_h^i, a_h^i)\bphi(s_h^i, a_h^i)^\top)\wb_h(f)$ yields
\begin{align}
  \|\hat{\wb}(f) - \wb_h(f)\|_{\bSigma_{k,h}}
  &= \nbr{-\lambda\wb_h(f) + \sum_{i=1}^{k-1} \bar{\sigma}_{i,h}^{-2}\bphi(s_h^i, a_h^i)\Big(f(s_{h+1}^i) - [\PP_h f](s_h^i, a_h^i)\Big)}_{\bSigma_{k,h}^{-1}} \nonumber \\
  &\le \sqrt{\lambda}\|\wb_h(f)\|_2 + \nbr{\sum_{i=1}^{k-1} \bar{\sigma}_{i,h}^{-2}\bphi(s_h^i, a_h^i)\Big(f(s_{h+1}^i) - [\PP_h f](s_h^i, a_h^i)\Big)}_{\bSigma_{k,h}^{-1}}, \label{eq:ridge-error-decomp}
\end{align}
where the inequality applies the triangle inequality and $\|\lambda\wb_h(f)\|_{\bSigma_{k,h}^{-1}} \le \sqrt{\lambda}\|\wb_h(f)\|_2$ (since $\bSigma_{k,h} \succeq \lambda\Ib$). By Proposition~\ref{prop:linear-moments}, the true parameters $\wb_{k,h}^* := \wb_h(V_{k,h+1})$, $\check{\wb}_{k,h}^* := \wb_h(\check{V}_{k,h+1})$, and $\tilde{\wb}_{k,h}^* := \wb_h(V_{k,h+1}^2)$ satisfy $\|\wb_{k,h}^*\|_2 \le H\sqrt{d}$, $\|\check{\wb}_{k,h}^*\|_2 \le H\sqrt{d}$, and $\|\tilde{\wb}_{k,h}^*\|_2 \le H^2\sqrt{d}$. When $k = 1$, the sum over $i$ in \eqref{eq:ridge-error-decomp} is empty, so \eqref{eq:ridge-cs}--\eqref{eq:ridge-error-decomp} immediately yield $\textup{\textbf{(P1)}}_{1,h}$ since $H\sqrt{d\lambda} \le \beta = \bar{\beta}$ and $H^2\sqrt{d\lambda} \le \tilde{\beta}$ by \eqref{eq:def-radii}. Now suppose $k \ge 2$. We bound \eqref{eq:ridge-error-decomp} for each of the three estimators in turn.

\textbf{Case (a): Concentration of $\hat{\wb}_{k,h}$.} Decompose $V_{k,h+1} = V_{h+1}^* + \Delta V_{k,h+1}$ as in \eqref{eq:three-way-decomp}, with $\Delta V_{k,h+1} \in \Delta\cV_{h+1}$. By \eqref{eq:ridge-error-decomp} and the triangle inequality, we have
\begin{align}
  \|\hat{\wb}_{k,h} - \wb_{k,h}^*\|_{\bSigma_{k,h}}
  &\le H\sqrt{d\lambda} + \underbrace{\nbr{\sum_{i=1}^{k-1} \bar{\sigma}_{i,h}^{-2}\bphi(s_h^i, a_h^i)\Big(V_{h+1}^*(s_{h+1}^i) - [\PP_h V_{h+1}^*](s_h^i, a_h^i)\Big)}_{\bSigma_{k,h}^{-1}}}_{=: J_1} \nonumber \\
  &\quad + \underbrace{\nbr{\sum_{i=1}^{k-1} \bar{\sigma}_{i,h}^{-2}\bphi(s_h^i, a_h^i)\Big(\Delta V_{k,h+1}(s_{h+1}^i) - [\PP_h \Delta V_{k,h+1}](s_h^i, a_h^i)\Big)}_{\bSigma_{k,h}^{-1}}}_{=: J_2}. \label{eq:w-hat-J1-J2}
\end{align}
For $J_1$, by $\textup{\textbf{(P2)}}_{i,h}$ at each $i \in \{1, \dots, k-1\}$, we have $[\VV_h V_{h+1}^*](s_h^i, a_h^i) \le \bar{\sigma}_{i,h}^2$, so the indicator in $\eta_{i,h}(V_{h+1}^*, 1)$ equals $1$ for all $i < k$, and \eqref{eq:Econc-Vstar} gives $J_1 \le \beta_0$.

For $J_2$, let $\tilde{V} \in \cN_{\epsilon_1}(\Delta\cV_{h+1})$ satisfy $\|\Delta V_{k,h+1} - \tilde{V}\|_\infty \le \epsilon_1$. Since $\|\Delta V_{k,h+1}\|_\infty \le H$ and $\|\tilde{V}\|_\infty \le H$, for every $i \in \{1, \dots, k-1\}$ we have
\begin{align}
  &[\VV_h \tilde{V}](s_h^i, a_h^i) - [\VV_h \Delta V_{k,h+1}](s_h^i, a_h^i) \nonumber \\
  &\quad = \big[\PP_h\big(\tilde{V}^2 - (\Delta V_{k,h+1})^2\big)\big](s_h^i, a_h^i) + \big([\PP_h \Delta V_{k,h+1}](s_h^i, a_h^i)\big)^2 - \big([\PP_h \tilde{V}](s_h^i, a_h^i)\big)^2 \nonumber \\
  &\quad \le 4 H \|\Delta V_{k,h+1} - \tilde{V}\|_\infty \le 4 H \epsilon_1 \le \frac{\alpha^2}{2 d^3 H}, \label{eq:var-net-diff}
\end{align}
where the first inequality uses $|a^2 - b^2| \le 2H|a - b|$ for $a, b \in [-H, H]$ and the last inequality uses \eqref{eq:def-eps1}. Consequently, for every $i \in \{1, \dots, k-1\}$,
\[
  [\VV_h \tilde{V}](s_h^i, a_h^i)
  \stackrel{\textup{(i)}}{\le} [\VV_h \Delta V_{k,h+1}](s_h^i, a_h^i) + \frac{\alpha^2}{2 d^3 H}
  \stackrel{\textup{(ii)}}{\le} \frac{D_{i,h} + \alpha^2}{2 d^3 H}
  \stackrel{\textup{(iii)}}{\le} \frac{\bar{\sigma}_{i,h}^2}{d^3 H},
\]
where (i) follows from \eqref{eq:var-net-diff}, (ii) applies $\textup{\textbf{(P3)}}_{i,h}$ (\eqref{eq:P3-opt} at $j = k$), and (iii) uses $\bar{\sigma}_{i,h}^2 \ge D_{i,h} + \alpha^2$ from \eqref{eq:def-sigma-kh}--\eqref{eq:def-sigma-bar}. Hence the indicator in $\eta_{i,h}(\tilde{V}, 1/(d^3 H))$ equals $1$ for all $i < k$, and the triangle inequality yields
\begin{align}
  J_2 &\le \nbr{\sum_{i=1}^{k-1} \xb_{i,h} \eta_{i,h}\!\rbr{\tilde{V}, \frac{1}{d^3 H}}}_{\bSigma_{k,h}^{-1}} \nonumber \\
  &\quad + \nbr{\sum_{i=1}^{k-1} \bar{\sigma}_{i,h}^{-2}\bphi(s_h^i, a_h^i)\Big((\Delta V_{k,h+1} - \tilde{V})(s_{h+1}^i) - [\PP_h(\Delta V_{k,h+1} - \tilde{V})](s_h^i, a_h^i)\Big)}_{\bSigma_{k,h}^{-1}} \nonumber \\
  &\le \beta_0 + \frac{2 K \epsilon_1}{\alpha^2\sqrt{\lambda}} = \beta_0 + \frac{1}{8 d^3 H^2} \le \beta_0 + 1, \label{eq:J2-bound}
\end{align}
where the second inequality uses \eqref{eq:Econc-net-opt}, $\bar{\sigma}_{i,h} \ge \alpha$, $\|\bSigma_{k,h}^{-1/2}\bphi(s_h^i, a_h^i)\|_2 \le \lambda^{-1/2}$, and $\|\Delta V_{k,h+1} - \tilde{V}\|_\infty \le \epsilon_1$, and the equality substitutes \eqref{eq:def-eps1}. Combining $J_1 \le \beta_0$ and \eqref{eq:J2-bound} with \eqref{eq:w-hat-J1-J2} and \eqref{eq:ridge-cs} proves \eqref{eq:P1-opt}.

\textbf{Case (b): Concentration of $\check{\wb}_{k,h}$.} Symmetrically, decompose $\check{V}_{k,h+1} = V_{h+1}^* - \Delta\check{V}_{k,h+1}$ as in \eqref{eq:three-way-decomp}, with $\Delta\check{V}_{k,h+1} \in \Delta\check{\cV}_{h+1}$, and choose $\tilde{U} \in \cN_{\epsilon_1}(\Delta\check{\cV}_{h+1})$ satisfying $\|\Delta\check{V}_{k,h+1} - \tilde{U}\|_\infty \le \epsilon_1$. For every $i < k$, the same Lipschitz bound as in \eqref{eq:var-net-diff} combined with \eqref{eq:P3-pess} of $\textup{\textbf{(P3)}}_{i,h}$ (at $j = k$) gives $[\VV_h \tilde{U}](s_h^i, a_h^i) \le (D_{i,h} + \alpha^2)/(2 d^3 H) \le \bar{\sigma}_{i,h}^2 / (d^3 H)$. Repeating \eqref{eq:w-hat-J1-J2}--\eqref{eq:J2-bound} via \eqref{eq:Econc-Vstar} and \eqref{eq:Econc-net-pess} yields $\|\check{\wb}_{k,h} - \check{\wb}_{k,h}^*\|_{\bSigma_{k,h}} \le H\sqrt{d\lambda} + 2\beta_0 + 1 = \bar{\beta}$, proving \eqref{eq:P1-pess}.

\textbf{Case (c): Concentration of $\tilde{\wb}_{k,h}$.} Decompose $V_{k,h+1}^2 = (V_{h+1}^*)^2 + \Delta(V_{k,h+1}^2)$ as in \eqref{eq:three-way-decomp}, with $\Delta(V_{k,h+1}^2) \in \Delta(\cV_{h+1}^2)$. By \eqref{eq:ridge-error-decomp} and the triangle inequality, we have
\begin{align}
  \|\tilde{\wb}_{k,h} - \tilde{\wb}_{k,h}^*\|_{\bSigma_{k,h}}
  &\le H^2\sqrt{d\lambda} + \underbrace{\nbr{\sum_{i=1}^{k-1} \bar{\sigma}_{i,h}^{-2}\bphi(s_h^i, a_h^i)\Big((V_{h+1}^*)^2(s_{h+1}^i) - [\PP_h (V_{h+1}^*)^2](s_h^i, a_h^i)\Big)}_{\bSigma_{k,h}^{-1}}}_{=: \tilde{J}_1} \nonumber \\
  &\quad + \underbrace{\nbr{\sum_{i=1}^{k-1} \bar{\sigma}_{i,h}^{-2}\bphi(s_h^i, a_h^i)\Big(\Delta(V_{k,h+1}^2)(s_{h+1}^i) - [\PP_h \Delta(V_{k,h+1}^2)](s_h^i, a_h^i)\Big)}_{\bSigma_{k,h}^{-1}}}_{=: \tilde{J}_2}. \label{eq:w-tilde-J1-J2}
\end{align}
For $\tilde{J}_1$, Corollary~\ref{cor:var-squared} and $\textup{\textbf{(P2)}}_{i,h}$ imply $[\VV_h (V_{h+1}^*)^2](s_h^i, a_h^i) \le 4 H^2 [\VV_h V_{h+1}^*](s_h^i, a_h^i) \le 4 H^2 \bar{\sigma}_{i,h}^2$ for all $i < k$, so the indicator in $\eta_{i,h}((V_{h+1}^*)^2, 4 H^2)$ equals $1$ and \eqref{eq:Econc-Vstar-sq} gives $\tilde{J}_1 \le 2 H \beta_0$. For $\tilde{J}_2$, let $\tilde{W} \in \cN_{\epsilon_2}(\Delta(\cV_{h+1}^2))$ satisfy $\|\Delta(V_{k,h+1}^2) - \tilde{W}\|_\infty \le \epsilon_2$. Since $\|\Delta(V_{k,h+1}^2)\|_\infty \le H^2$ and $\|\tilde{W}\|_\infty \le H^2$, for every $i \in \{1, \dots, k-1\}$ we have
\[
  [\VV_h \tilde{W}](s_h^i, a_h^i)
  \stackrel{\textup{(i)}}{\le} \big[\VV_h \Delta(V_{k,h+1}^2)\big](s_h^i, a_h^i) + 4 H^2 \epsilon_2
  \stackrel{\textup{(ii)}}{\le} \frac{H^2 D_{i,h}}{2 d^3 H} + \frac{\alpha^2}{2 d^3 K}
  \stackrel{\textup{(iii)}}{\le} \frac{H^2 \bar{\sigma}_{i,h}^2}{d^3 H},
\]
where (i) uses $|a^2 - b^2| \le 2 H^2 |a - b|$ for $a, b \in [-H^2, H^2]$, (ii) applies \eqref{eq:P3-sq} of $\textup{\textbf{(P3)}}_{i,h}$ (at $j = k$) together with $4 H^2 \epsilon_2 = \alpha^2 / (2 d^3 K)$ from \eqref{eq:def-eps2}, and (iii) holds since $H, K \ge 1$ and $\bar{\sigma}_{i,h}^2 \ge D_{i,h} + \alpha^2$. Thus the indicator in $\eta_{i,h}(\tilde{W}, H^2/(d^3 H))$ equals $1$ for all $i < k$, and the triangle inequality with \eqref{eq:Econc-net-sq} gives
\[
  \tilde{J}_2 \le H\beta_0 + \frac{2 K \epsilon_2}{\alpha^2\sqrt{\lambda}} = H\beta_0 + \frac{1}{4 d^3 H} \le H\beta_0 + H,
\]
where the equality uses \eqref{eq:def-eps2}. Substituting $\tilde{J}_1 \le 2 H\beta_0$ and $\tilde{J}_2 \le H\beta_0 + H$ into \eqref{eq:w-tilde-J1-J2} yields $\|\tilde{\wb}_{k,h} - \tilde{\wb}_{k,h}^*\|_{\bSigma_{k,h}} \le H^2\sqrt{d\lambda} + 3 H\beta_0 + H \le 2 H\beta = \tilde{\beta}$, which together with \eqref{eq:ridge-cs} establishes \eqref{eq:P1-sq}.

\paragraph{Step 2: Proof of $\textup{(P2)}_{k,h}$.}
From \eqref{eq:def-emp-var}, we have
\begin{align}
  &\abr{[\bar{\VV}_{k,h} V_{k,h+1}](s_h^k, a_h^k) - [\VV_h V_{k,h+1}](s_h^k, a_h^k)} \nonumber \\
  &\quad \le \abr{\sbr{\tilde{\wb}_{k,h}^\top \bphi(s_h^k, a_h^k)}_{[0,H^2]} - [\PP_h V_{k,h+1}^2](s_h^k, a_h^k)} + \abr{\sbr{\hat{\wb}_{k,h}^\top \bphi(s_h^k, a_h^k)}_{[0,H]}^2 - \big([\PP_h V_{k,h+1}](s_h^k, a_h^k)\big)^2} \nonumber \\
  &\quad \le \min\cbr{\tilde{\beta}\|\bphi(s_h^k, a_h^k)\|_{\bSigma_{k,h}^{-1}}, \, H^2} + \min\cbr{2 H \bar{\beta}\|\bphi(s_h^k, a_h^k)\|_{\bSigma_{k,h}^{-1}}, \, H^2} = E_{k,h}, \label{eq:P2-step1}
\end{align}
where the first inequality uses the $1$-Lipschitz property of $[\cdot]_{[0,H^2]}$ together with the triangle inequality, the second inequality uses $[\PP_h V_{k,h+1}^2](s_h^k, a_h^k) \in [0, H^2]$, $[\PP_h V_{k,h+1}](s_h^k, a_h^k) \in [0, H]$, $|a^2 - b^2| \le 2 H |a - b|$ for $a, b \in [0,H]$, \eqref{eq:P1-opt}, \eqref{eq:P1-sq}, and $\bar{\beta} = \beta$, and the equality uses \eqref{eq:def-Ekh}. This proves \eqref{eq:P2-Vk}. Next, we bound the variance gap between $V_{k,h+1}$ and $V_{h+1}^*$:
\begin{align}
  &\abr{[\VV_h V_{k,h+1}](s_h^k, a_h^k) - [\VV_h V_{h+1}^*](s_h^k, a_h^k)} \nonumber \\
  &\quad \le \abr{\big[\PP_h\big(V_{k,h+1}^2 - (V_{h+1}^*)^2\big)\big](s_h^k, a_h^k)} + \abr{\big([\PP_h V_{k,h+1}](s_h^k, a_h^k)\big)^2 - \big([\PP_h V_{h+1}^*](s_h^k, a_h^k)\big)^2} \nonumber \\
  &\quad \le 4 H [\PP_h(V_{k,h+1} - V_{h+1}^*)](s_h^k, a_h^k) \le 4 H [\PP_h(V_{k,h+1} - \check{V}_{k,h+1})](s_h^k, a_h^k) \nonumber \\
  &\quad \le 4 H \rbr{(\hat{\wb}_{k,h} - \check{\wb}_{k,h})^\top \bphi(s_h^k, a_h^k) + 2\bar{\beta}\|\bphi(s_h^k, a_h^k)\|_{\bSigma_{k,h}^{-1}}}_+, \label{eq:P2-step2}
\end{align}
where the second and third inequalities follow from \eqref{eq:outer-ind-hyp} at $m = k$ ($0 \le \check{V}_{k,h+1} \le V_{h+1}^* \le V_{k,h+1} \le H$), and the last inequality uses \eqref{eq:P1-opt}, \eqref{eq:P1-pess}, and $\bar{\beta} = \beta$. Since both $[\VV_h V_{k,h+1}](s_h^k, a_h^k)$ and $[\VV_h V_{h+1}^*](s_h^k, a_h^k)$ lie in $[0, H^2]$, their absolute difference is also at most $H^2 \le d^3 H^3$, so \eqref{eq:P2-step2} and \eqref{eq:def-Dkh} imply $|[\VV_h V_{k,h+1}](s_h^k, a_h^k) - [\VV_h V_{h+1}^*](s_h^k, a_h^k)| \le D_{k,h}/(d^3 H) \le D_{k,h}$. Combining this bound with \eqref{eq:P2-step1} via the triangle inequality proves \eqref{eq:P2-Vstar}.

\paragraph{Step 3: Proof of $\textup{(P3)}_{k,h}$.}
Fix any $j \in \{k, \dots, K\}$. For all $s' \in \cS$, we have
\begin{align}
  0 \le V_{j,h+1}(s') - V_{h+1}^*(s') &\le V_{k,h+1}(s') - \check{V}_{k,h+1}(s') \le H, \label{eq:mono-opt} \\
  0 \le V_{h+1}^*(s') - \check{V}_{j,h+1}(s') &\le V_{k,h+1}(s') - \check{V}_{k,h+1}(s') \le H, \label{eq:mono-pess} \\
  0 \le V_{j,h+1}^2(s') - (V_{h+1}^*)^2(s') &\le \min\cbr{2 H \big(V_{k,h+1}(s') - \check{V}_{k,h+1}(s')\big), \, H^2}, \label{eq:mono-sq}
\end{align}
where \eqref{eq:mono-opt} and \eqref{eq:mono-pess} follow from \eqref{eq:outer-ind-hyp} at $m = j$ together with the pointwise monotonicity $V_{j,h+1} \le V_{k,h+1}$ and $\check{V}_{j,h+1} \ge \check{V}_{k,h+1}$ from Lines~14--18 of Algorithm~\ref{alg:va-lsvi-ucbpp}, and \eqref{eq:mono-sq} uses $V_{j,h+1}^2 - (V_{h+1}^*)^2 = (V_{j,h+1} + V_{h+1}^*)(V_{j,h+1} - V_{h+1}^*) \le \min\{2 H(V_{k,h+1} - \check{V}_{k,h+1}), H^2\}$. For the first-moment residuals, we obtain
\begin{align*}
  &\max\cbr{\big[\VV_h(V_{j,h+1} - V_{h+1}^*)\big](s_h^k, a_h^k), \, \big[\VV_h(V_{h+1}^* - \check{V}_{j,h+1})\big](s_h^k, a_h^k)} \\
  &\quad \le \min\cbr{H [\PP_h(V_{k,h+1} - \check{V}_{k,h+1})](s_h^k, a_h^k), \, \frac{H^2}{2}} \le \frac{D_{k,h}}{2 d^3 H},
\end{align*}
where the first inequality uses $\Var(Z) \le \min\{\EE[Z^2], B^2/4\} \le \min\{B\EE[Z], B^2/2\}$ for any random variable $Z \in [0, B]$ together with \eqref{eq:mono-opt}--\eqref{eq:mono-pess} ($B = H$), and the second inequality follows from \eqref{eq:P1-opt}--\eqref{eq:P1-pess}, $\bar{\beta} = \beta$, and the definition of $D_{k,h}$ in \eqref{eq:def-Dkh}. Similarly, for the squared residual, we have
\[
  \big[\VV_h\big(V_{j,h+1}^2 - (V_{h+1}^*)^2\big)\big](s_h^k, a_h^k)
  \le \min\cbr{2 H^3 [\PP_h(V_{k,h+1} - \check{V}_{k,h+1})](s_h^k, a_h^k), \, \frac{H^4}{2}}
  \le \frac{H^2 D_{k,h}}{2 d^3 H},
\]
where the first inequality applies $\Var(Z) \le \min\{B\EE[Z], B^2/2\}$ with $B = H^2$ to \eqref{eq:mono-sq}, and the second inequality again uses \eqref{eq:P1-opt}--\eqref{eq:P1-pess}, $\bar{\beta} = \beta$, and \eqref{eq:def-Dkh}. This proves \eqref{eq:P3-opt}--\eqref{eq:P3-sq} and completes the inner induction over $k = 1, \dots, K$.

\paragraph{Step 4: Proof of $\textup{(P4)}_{k,h}$ for $k \in [K]$.}
For all $k \in [K]$ and $(s,a) \in \cS \times \cA$, we have
\begin{align*}
  r_h(s,a) + \hat{\wb}_{k,h}^\top \bphi(s,a) + \beta\|\bphi(s,a)\|_{\bSigma_{k,h}^{-1}}
  &\ge r_h(s,a) + [\PP_h V_{k,h+1}](s,a) \ge Q_h^*(s,a), \\
  r_h(s,a) + \check{\wb}_{k,h}^\top \bphi(s,a) - \bar{\beta}\|\bphi(s,a)\|_{\bSigma_{k,h}^{-1}}
  &\le r_h(s,a) + [\PP_h \check{V}_{k,h+1}](s,a) \le Q_h^*(s,a),
\end{align*}
where the first inequalities follow from \eqref{eq:P1-opt} and \eqref{eq:P1-pess}, and the second inequalities use \eqref{eq:outer-ind-hyp} and the Bellman optimality equation $Q_h^*(s,a) = r_h(s,a) + [\PP_h V_{h+1}^*](s,a)$. Because $0 = \check{Q}_{0,h}(s,a) \le Q_h^*(s,a) \le Q_{0,h}(s,a) = H$, a direct induction on $k = 1, \dots, K$ via \eqref{eq:update-Q-opt}--\eqref{eq:update-Q-pess} and Lines~16--18 of Algorithm~\ref{alg:va-lsvi-ucbpp} establishes $\check{Q}_{k,h}(s,a) \le Q_h^*(s,a) \le Q_{k,h}(s,a)$ and $\check{V}_{k,h}(s) \le V_h^*(s) \le V_{k,h}(s)$ for all $k \in [K]$. This completes the outer induction over $h = H \to 1$.
\end{proof}

\section{Regret Analysis and Proof of Theorem~\ref{thm:main}}
\label{app:regret}

Throughout this appendix, we work with the following shorthand quantities along the learner's trajectory:
\begin{align}
  B_{k,h} &:= 4\min\cbr{\beta \|\bphi(s_h^k, a_h^k)\|_{\bSigma_{k,h}^{-1}}, \, H}, \label{eq:def-Bkh} \\
  \Vest &:= \sum_{k=1}^K \sum_{h=1}^H \sigma_{k,h}^2, \label{eq:def-Vest} \\
  S_\pi &:= \sum_{k=1}^K \sum_{h=1}^H [\PP_h(V_{k,h+1} - V_{h+1}^{\pi^k})](s_h^k, a_h^k), \label{eq:def-Spi} \\
  S_{\ul} &:= \sum_{k=1}^K \sum_{h=1}^H [\PP_h(V_{k,h+1} - \check{V}_{k,h+1})](s_h^k, a_h^k). \label{eq:def-Sul}
\end{align}
In the following subsections, we sequentially establish the elliptical bonus summation bound (Lemma~\ref{lem:bonus-sum}), the Freedman-type value difference bounds (Lemma~\ref{lem:freedman-value-diff}), the total estimated variance bound (Lemma~\ref{lem:total-est-var}), and the proof of Theorem~\ref{thm:main}.

\subsection{Elliptical Bonus Summation}

\begin{lemma}[Refined Elliptical Bonus Summation]
\label{lem:bonus-sum}
For every stage $h \in [H]$ and any parameters $\beta' > 0$, $C > 0$, the sequence of covariance matrices $\{\bSigma_{k,h}\}_{k=1}^K$ generated by Algorithm~\ref{alg:va-lsvi-ucbpp} satisfies
\begin{equation}
  \sum_{k=1}^K \min\cbr{\beta' \|\bphi(s_h^k, a_h^k)\|_{\bSigma_{k,h}^{-1}}, \, C} \le 2 C d \iota + 2 \beta' \gamma^2 d \iota + \beta' \sqrt{2 d \iota \sum_{k=1}^K \sigma_{k,h}^2}. \label{eq:bonus-sum-bound}
\end{equation}
In particular, the cumulative truncated bonus $B_{k,h}$ in \eqref{eq:def-Bkh} satisfies
\begin{equation}
  \sum_{k=1}^K \sum_{h=1}^H B_{k,h} \le 8 d H^2 \iota + 8 \beta \gamma^2 d H \iota + 4\beta\sqrt{2 d H \iota \Vest}. \label{eq:sum-Bkh-Vest}
\end{equation}
\end{lemma}

\begin{proof}
Recall that $\xb_{k,h} = \bar{\sigma}_{k,h}^{-1}\bphi(s_h^k, a_h^k)$, so that $\bSigma_{k+1,h} = \bSigma_{k,h} + \xb_{k,h}\xb_{k,h}^\top$ and $\|\xb_{k,h}\|_2 \le \alpha^{-1}$. By Lemma~\ref{lem:elliptical-potential} and \eqref{eq:def-iotas}, we have
\begin{equation}
  \sum_{k=1}^K \min\cbr{1, \, \|\xb_{k,h}\|_{\bSigma_{k,h}^{-1}}^2} \le 2 d \ln\!\rbr{1 + \frac{K}{d \lambda \alpha^2}} \le 2 d \iota. \label{eq:elliptical-sum}
\end{equation}
Partition $[K]$ into three disjoint index sets:
\begin{align*}
  \cK_1 &:= \cbr{k \in [K] : \|\xb_{k,h}\|_{\bSigma_{k,h}^{-1}} \ge 1}, \qquad
  \cK_2 := \cbr{k \notin \cK_1 : \bar{\sigma}_{k,h} = \gamma \|\bphi(s_h^k, a_h^k)\|_{\bSigma_{k,h}^{-1}}^{1/2}}, \\
  \cK_3 &:= [K] \setminus (\cK_1 \cup \cK_2).
\end{align*}
Splitting the sum in \eqref{eq:bonus-sum-bound} across $\cK_1$, $\cK_2$, and $\cK_3$ yields
\begin{align*}
  &\sum_{k=1}^K \min\cbr{\beta' \|\bphi(s_h^k, a_h^k)\|_{\bSigma_{k,h}^{-1}}, \, C} \\
  &\quad \le \sum_{k \in \cK_1} C + \beta' \gamma^2 \sum_{k \in \cK_2} \min\cbr{1, \, \|\xb_{k,h}\|_{\bSigma_{k,h}^{-1}}^2} + \beta' \sum_{k \in \cK_3} \sigma_{k,h} \min\cbr{1, \, \|\xb_{k,h}\|_{\bSigma_{k,h}^{-1}}} \\
  &\quad \le 2 C d \iota + 2 \beta' \gamma^2 d \iota + \beta' \sqrt{\sum_{k=1}^K \sigma_{k,h}^2} \sqrt{\sum_{k=1}^K \min\cbr{1, \, \|\xb_{k,h}\|_{\bSigma_{k,h}^{-1}}^2}} \\
  &\quad \le 2 C d \iota + 2 \beta' \gamma^2 d \iota + \beta' \sqrt{2 d \iota \sum_{k=1}^K \sigma_{k,h}^2},
\end{align*}
where the first inequality uses $\|\bphi(s_h^k, a_h^k)\|_{\bSigma_{k,h}^{-1}} = \gamma^2 \|\xb_{k,h}\|_{\bSigma_{k,h}^{-1}}^2 = \gamma^2 \min\{1, \|\xb_{k,h}\|_{\bSigma_{k,h}^{-1}}^2\}$ for $k \in \cK_2$ and $\|\bphi(s_h^k, a_h^k)\|_{\bSigma_{k,h}^{-1}} = \sigma_{k,h} \|\xb_{k,h}\|_{\bSigma_{k,h}^{-1}} = \sigma_{k,h}\min\{1, \|\xb_{k,h}\|_{\bSigma_{k,h}^{-1}}\}$ for $k \in \cK_3$ (since $\bar{\sigma}_{k,h} = \sigma_{k,h}$ on $\cK_3$ by \eqref{eq:def-sigma-bar}), the second inequality follows from $|\cK_1| \le \sum_{k\in\cK_1}\min\{1, \|\xb_{k,h}\|_{\bSigma_{k,h}^{-1}}^2\} \le 2 d \iota$ by \eqref{eq:elliptical-sum} and the Cauchy--Schwarz inequality, and the last inequality applies \eqref{eq:elliptical-sum} once more. Finally, summing \eqref{eq:bonus-sum-bound} with $(\beta', C) = (4\beta, 4H)$ over $h \in [H]$ and applying $\sum_{h=1}^H \sqrt{\sum_{k=1}^K \sigma_{k,h}^2} \le \sqrt{H \Vest}$ from \eqref{eq:def-Vest} proves \eqref{eq:sum-Bkh-Vest}.
\end{proof}

\subsection{Freedman-Type Value Difference Bounds}

\begin{lemma}[Value Difference Bounds]
\label{lem:freedman-value-diff}
There exists an event $\EFreed$ with $\PP(\EFreed) \ge 1 - 2\delta$ such that on $\Econc \cap \EFreed$, the quantities $B_{k,h}$, $S_\pi$, and $S_{\ul}$ defined in \eqref{eq:def-Bkh}, \eqref{eq:def-Spi}, and \eqref{eq:def-Sul} satisfy
\begin{align}
  \Regret(K) &\le 2 \sum_{k=1}^K \sum_{h=1}^H B_{k,h} + 4 H^2 \ln(H/\delta), \label{eq:freed-regret} \\
  S_\pi &\le 2 H \sum_{k=1}^K \sum_{h=1}^H B_{k,h} + 4 H^3 \ln(H/\delta), \label{eq:freed-S-pi} \\
  S_{\ul} &\le 4 H \sum_{k=1}^K \sum_{h=1}^H B_{k,h} + 4 H^3 \ln(H/\delta). \label{eq:freed-S-ul}
\end{align}
\end{lemma}

\begin{proof}
For each episode $k \in [K]$ and stage $h \in [H]$, define the truncated value differences
\[
  \tilde{\Delta}_{k,h+1}^\pi(s) := \sbr{V_{k,h+1}(s) - V_{h+1}^{\pi^k}(s)}_{[0,H]}, \qquad \tilde{\Delta}_{k,h+1}^{\ul}(s) := \sbr{V_{k,h+1}(s) - \check{V}_{k,h+1}(s)}_{[0,H]},
\]
and the corresponding martingale differences:
\[
  M_{k,h}^\pi := [\PP_h \tilde{\Delta}_{k,h+1}^\pi](s_h^k, a_h^k) - \tilde{\Delta}_{k,h+1}^\pi(s_{h+1}^k), \qquad M_{k,h}^{\ul} := [\PP_h \tilde{\Delta}_{k,h+1}^{\ul}](s_h^k, a_h^k) - \tilde{\Delta}_{k,h+1}^{\ul}(s_{h+1}^k).
\]
Because $V_{k,h+1}$, $\check{V}_{k,h+1}$, and $\pi^k$ are fixed prior to episode $k$ and $\tilde{\Delta}_{k,h+1}^\pi(s'), \tilde{\Delta}_{k,h+1}^{\ul}(s') \in [0,H]$ unconditionally, we have $|M_{k,h}^\pi| \le H$, $\EE[M_{k,h}^\pi \mid \cF_{k,h}] = 0$, and the conditional variance is self-bounded by
\[
  \EE\big[(M_{k,h}^\pi)^2 \mid \cF_{k,h}\big] = [\VV_h \tilde{\Delta}_{k,h+1}^\pi](s_h^k, a_h^k) \le \big[\PP_h (\tilde{\Delta}_{k,h+1}^\pi)^2\big](s_h^k, a_h^k) \le H [\PP_h \tilde{\Delta}_{k,h+1}^\pi](s_h^k, a_h^k),
\]
with the identical bound holding for $M_{k,h}^{\ul}$. For each starting stage $h \in [H]$, applying Lemma~\ref{lem:freedman} to $\{M_{k,h'}^\pi\}_{k\in[K], h'\ge h}$ and $\{M_{k,h'}^{\ul}\}_{k\in[K], h'\ge h}$ with range $B = H$, step size $\eta = 1/(2 H^2) \in (0, 1/H]$, and failure probability $\delta/H$, and taking a union bound over $h \in [H]$, we obtain an event $\EFreed$ with $\PP(\EFreed) \ge 1 - 2\delta$ on which for all $h \in [H]$:
\begin{align}
  \sum_{k=1}^K \sum_{h'=h}^H M_{k,h'}^\pi &\le \frac{1}{2 H} \sum_{k=1}^K \sum_{h'=h}^H [\PP_{h'} \tilde{\Delta}_{k,h'+1}^\pi](s_{h'}^k, a_{h'}^k) + 2 H^2 \ln(H/\delta), \label{eq:M-pi-tail} \\
  \sum_{k=1}^K \sum_{h'=h}^H M_{k,h'}^{\ul} &\le \frac{1}{2 H} \sum_{k=1}^K \sum_{h'=h}^H [\PP_{h'} \tilde{\Delta}_{k,h'+1}^{\ul}](s_{h'}^k, a_{h'}^k) + 2 H^2 \ln(H/\delta). \label{eq:M-ul-tail}
\end{align}

Now condition on $\Econc \cap \EFreed$. By property~$\textup{\textbf{(P4)}}_{k,h}$ of Lemma~\ref{lem:double-induction}, we have $0 \le \check{V}_{k,h} \le V_h^* \le V_{k,h} \le H$ and $0 \le V_h^{\pi^k} \le V_h^* \le V_{k,h} \le H$, so $\tilde{\Delta}_{k,h}^\pi = V_{k,h} - V_h^{\pi^k}$ and $\tilde{\Delta}_{k,h}^{\ul} = V_{k,h} - \check{V}_{k,h}$ for all $(k,h) \in [K] \times [H]$. Let $\klast \le k$ denote the most recent episode at which Line~4 of Algorithm~\ref{alg:va-lsvi-ucbpp} was triggered prior to or at episode $k$. For every $k \in [K]$ and $h \in [H]$, we first bound $Q_{\klast,h}(s_h^k, a_h^k) - Q_h^{\pi^k}(s_h^k, a_h^k)$:
\begin{align*}
  Q_{\klast,h}(s_h^k, a_h^k) - Q_h^{\pi^k}(s_h^k, a_h^k)
  &\le r_h(s_h^k, a_h^k) + \hat{\wb}_{\klast,h}^\top \bphi(s_h^k, a_h^k) + \beta\|\bphi(s_h^k, a_h^k)\|_{\bSigma_{\klast,h}^{-1}} - Q_h^{\pi^k}(s_h^k, a_h^k) \\
  &= \big(\hat{\wb}_{\klast,h}^\top \bphi(s_h^k, a_h^k) - [\PP_h V_{\klast,h+1}](s_h^k, a_h^k)\big) \\
  &\quad + [\PP_h \tilde{\Delta}_{k,h+1}^\pi](s_h^k, a_h^k) + \beta\|\bphi(s_h^k, a_h^k)\|_{\bSigma_{\klast,h}^{-1}} \\
  &\le [\PP_h \tilde{\Delta}_{k,h+1}^\pi](s_h^k, a_h^k) + 2\beta\|\bphi(s_h^k, a_h^k)\|_{\bSigma_{\klast,h}^{-1}},
\end{align*}
where the first inequality follows from \eqref{eq:update-Q-opt}, the equality uses $Q_h^{\pi^k}(s_h^k, a_h^k) = r_h(s_h^k, a_h^k) + [\PP_h V_{h+1}^{\pi^k}](s_h^k, a_h^k)$ and $V_{\klast,h+1} = V_{k,h+1}$, and the second inequality applies \eqref{eq:P1-opt} of $\textup{\textbf{(P1)}}_{\klast,h}$. Since $\tilde{\Delta}_{k,h}^\pi(s_h^k) = Q_{\klast,h}(s_h^k, a_h^k) - Q_h^{\pi^k}(s_h^k, a_h^k) \le H$, we further have
\begin{align}
  \tilde{\Delta}_{k,h}^\pi(s_h^k)
  &\le \min\cbr{H, \, [\PP_h \tilde{\Delta}_{k,h+1}^\pi](s_h^k, a_h^k) + 2\beta\|\bphi(s_h^k, a_h^k)\|_{\bSigma_{\klast,h}^{-1}}} \nonumber \\
  &\le [\PP_h \tilde{\Delta}_{k,h+1}^\pi](s_h^k, a_h^k) + 2\min\cbr{\beta\|\bphi(s_h^k, a_h^k)\|_{\bSigma_{\klast,h}^{-1}}, \, H} \nonumber \\
  &\le [\PP_h \tilde{\Delta}_{k,h+1}^\pi](s_h^k, a_h^k) + B_{k,h} = \tilde{\Delta}_{k,h+1}^\pi(s_{h+1}^k) + M_{k,h}^\pi + B_{k,h}, \label{eq:one-step-pi}
\end{align}
where the second inequality uses $[\PP_h \tilde{\Delta}_{k,h+1}^\pi](s_h^k, a_h^k) \ge 0$, and the third inequality uses \eqref{eq:def-Bkh} and Lemma~\ref{lem:det-ratio}, which gives
\[
  \|\bphi(s_h^k, a_h^k)\|_{\bSigma_{\klast,h}^{-1}} \le \sqrt{2}\|\bphi(s_h^k, a_h^k)\|_{\bSigma_{k,h}^{-1}} \le 2\|\bphi(s_h^k, a_h^k)\|_{\bSigma_{k,h}^{-1}}
\]
since $\bSigma_{k,h} \succeq \bSigma_{\klast,h}$ and $\det(\bSigma_{k,h}) < 2\det(\bSigma_{\klast,h})$. Telescoping \eqref{eq:one-step-pi} over $h' = h, \dots, H$ (with $\tilde{\Delta}_{k,H+1}^\pi = 0$) and summing over $k \in [K]$ yields, for every $h \in [H]$:
\begin{equation}
  \sum_{k=1}^K \tilde{\Delta}_{k,h}^\pi(s_h^k) \le \sum_{k=1}^K \sum_{h'=h}^H B_{k,h'} + \sum_{k=1}^K \sum_{h'=h}^H M_{k,h'}^\pi \le \sum_{k=1}^K \sum_{h=1}^H B_{k,h} + \frac{1}{2 H} S_\pi + 2 H^2 \ln(H/\delta), \label{eq:stage-h-pi}
\end{equation}
where the second inequality applies \eqref{eq:M-pi-tail} and $\sum_{k=1}^K \sum_{h'=h}^H [\PP_{h'} \tilde{\Delta}_{k,h'+1}^\pi](s_{h'}^k, a_{h'}^k) \le S_\pi$ from \eqref{eq:def-Spi}. Consequently, expanding $S_\pi$ from \eqref{eq:def-Spi} gives
\begin{align*}
  S_\pi
  &= \sum_{k=1}^K \sum_{h=1}^H \rbr{\tilde{\Delta}_{k,h+1}^\pi(s_{h+1}^k) + M_{k,h}^\pi}
  = \sum_{h=2}^H \sum_{k=1}^K \tilde{\Delta}_{k,h}^\pi(s_h^k) + \sum_{k=1}^K \sum_{h=1}^H M_{k,h}^\pi \\
  &\le (H - 1)\rbr{\sum_{k=1}^K \sum_{h=1}^H B_{k,h} + \frac{1}{2 H} S_\pi + 2 H^2 \ln(H/\delta)} + \frac{1}{2 H} S_\pi + 2 H^2 \ln(H/\delta) \\
  &\le H \sum_{k=1}^K \sum_{h=1}^H B_{k,h} + \frac{1}{2} S_\pi + 2 H^3 \ln(H/\delta),
\end{align*}
where the second equality shifts the stage index $h+1 \mapsto h$ using $\tilde{\Delta}_{k,H+1}^\pi = 0$, and the first inequality applies \eqref{eq:stage-h-pi} for each $h \in \{2, \dots, H\}$ and \eqref{eq:M-pi-tail} at $h = 1$. Rearranging proves \eqref{eq:freed-S-pi}. Using \eqref{eq:freed-S-pi}, we bound the cumulative regret as
\begin{align*}
  \Regret(K)
  &\le \sum_{k=1}^K \rbr{V_{k,1}(s_1^k) - V_1^{\pi^k}(s_1^k)} = \sum_{k=1}^K \tilde{\Delta}_{k,1}^\pi(s_1^k) \\
  &\le \sum_{k=1}^K \sum_{h=1}^H B_{k,h} + \frac{1}{2 H} S_\pi + 2 H^2 \ln(H/\delta)
  \le 2 \sum_{k=1}^K \sum_{h=1}^H B_{k,h} + 4 H^2 \ln(H/\delta),
\end{align*}
where the first inequality holds by optimism ($\textup{\textbf{(P4)}}_{k,1}$ of Lemma~\ref{lem:double-induction}), the second inequality applies \eqref{eq:stage-h-pi} at $h = 1$, and the third substitutes \eqref{eq:freed-S-pi}. This establishes \eqref{eq:freed-regret}. Finally, since $\check{V}_{k,h}(s_h^k) = \max_{a\in\cA}\check{Q}_{\klast,h}(s_h^k, a) \ge \check{Q}_{\klast,h}(s_h^k, a_h^k)$ and $\bar{\beta} = \beta$, \eqref{eq:update-Q-pess} and \eqref{eq:P1-pess} imply that $\tilde{\Delta}_{k,h}^{\ul}(s_h^k)$ satisfies \eqref{eq:one-step-pi} with $B_{k,h}$ replaced by $2 B_{k,h}$ and $M_{k,h}^\pi$ replaced by $M_{k,h}^{\ul}$. Repeating the same telescoping derivation with \eqref{eq:M-ul-tail} and \eqref{eq:def-Sul} yields \eqref{eq:freed-S-ul}.
\end{proof}

\subsection{Total Estimated Variance Bound and Proof of Theorem~\ref{thm:main}}

\begin{lemma}[Total Estimated Variance Bound]
\label{lem:total-est-var}
On the event $\Econc \cap \EFreed$, the total estimated variance $\Vest$ defined in \eqref{eq:def-Vest} satisfies
\begin{align}
  \Vest &\le 2\min\cbr{\Varkstar, \Varkpi} + \cO\!\rbr{d^7 H^7 \beta^2 \iota + d^7 H^7 \beta \iota^3} \nonumber \\
  &\le 2\min\cbr{\Varkstar, \Varkpi} + \cO\!\rbr{d^8 H^7 \iota^4 \iota_0 + d^7 H^7 \iota^4 \iota_0^2}. \label{eq:total-est-var-bound}
\end{align}
\end{lemma}

\begin{proof}
Condition on $\Econc \cap \EFreed$. Expanding $\Vest$ from \eqref{eq:def-sigma-kh} and \eqref{eq:def-Vest} gives
\begin{equation}
  \Vest = \underbrace{\sum_{k=1}^K \sum_{h=1}^H [\bar{\VV}_{k,h} V_{k,h+1}](s_h^k, a_h^k)}_{=: (\diamond)} + \sum_{k=1}^K \sum_{h=1}^H E_{k,h} + \sum_{k=1}^K \sum_{h=1}^H D_{k,h} + H^3, \label{eq:Vest-expand}
\end{equation}
where we used $\sum_{k=1}^K \sum_{h=1}^H \alpha^2 = K H (H^2/K) = H^3$. We bound $(\diamond)$ in terms of $\Varkstar$ and $\Varkpi$ separately. On the one hand, summing \eqref{eq:P2-Vstar} of Lemma~\ref{lem:double-induction} over $(k,h) \in [K] \times [H]$ gives
\begin{equation}
  (\diamond) \le \Varkstar + \sum_{k=1}^K \sum_{h=1}^H E_{k,h} + \sum_{k=1}^K \sum_{h=1}^H D_{k,h}. \label{eq:emp-var-star}
\end{equation}
On the other hand, relating $[\bar{\VV}_{k,h} V_{k,h+1}](s_h^k, a_h^k)$ to $[\VV_h V_{h+1}^{\pi^k}](s_h^k, a_h^k)$ yields
\begin{align}
  (\diamond)
  &\le \sum_{k=1}^K \sum_{h=1}^H [\VV_h V_{k,h+1}](s_h^k, a_h^k) + \sum_{k=1}^K \sum_{h=1}^H E_{k,h} \nonumber \\
  &= \Varkpi + \sum_{k=1}^K \sum_{h=1}^H \rbr{[\VV_h V_{k,h+1}](s_h^k, a_h^k) - [\VV_h V_{h+1}^{\pi^k}](s_h^k, a_h^k)} + \sum_{k=1}^K \sum_{h=1}^H E_{k,h} \nonumber \\
  &\le \Varkpi + \sum_{k=1}^K \sum_{h=1}^H \big[\PP_h\big(V_{k,h+1}^2 - (V_{h+1}^{\pi^k})^2\big)\big](s_h^k, a_h^k) + \sum_{k=1}^K \sum_{h=1}^H E_{k,h} \nonumber \\
  &\le \Varkpi + 2 H S_\pi + \sum_{k=1}^K \sum_{h=1}^H E_{k,h}, \label{eq:emp-var-pi}
\end{align}
where the first inequality follows from \eqref{eq:P2-Vk} of Lemma~\ref{lem:double-induction}, the second inequality holds since $[\PP_h V_{k,h+1}](s_h^k, a_h^k) \ge [\PP_h V_{h+1}^{\pi^k}](s_h^k, a_h^k) \ge 0$ by $\textup{\textbf{(P4)}}_{k,h+1}$, and the last inequality uses $V_{k,h+1}^2 - (V_{h+1}^{\pi^k})^2 \le 2 H(V_{k,h+1} - V_{h+1}^{\pi^k})$ (since $0 \le V_{h+1}^{\pi^k} \le V_{k,h+1} \le H$) together with the definition of $S_\pi$ in \eqref{eq:def-Spi}. Substituting the minimum of \eqref{eq:emp-var-star} and \eqref{eq:emp-var-pi} into \eqref{eq:Vest-expand} gives
\begin{equation}
  \Vest \le \min\cbr{\Varkstar, \Varkpi} + \underbrace{2\sum_{k=1}^K \sum_{h=1}^H E_{k,h}}_{=: I_1} + \underbrace{2\sum_{k=1}^K \sum_{h=1}^H D_{k,h}}_{=: I_2} + 2 H S_\pi + H^3. \label{eq:Vest-master-decomp}
\end{equation}

We now bound $I_1$ and $I_2$ in turn. For $I_1$, expanding $E_{k,h}$ from \eqref{eq:def-Ekh} gives
\begin{align}
  I_1 &= 2\sum_{h=1}^H \sum_{k=1}^K \min\cbr{2 H\beta\|\bphi(s_h^k, a_h^k)\|_{\bSigma_{k,h}^{-1}}, \, H^2} + 2\sum_{h=1}^H \sum_{k=1}^K \min\cbr{2 H\beta\|\bphi(s_h^k, a_h^k)\|_{\bSigma_{k,h}^{-1}}, \, H^2} \nonumber \\
  &\le 8 d H^3 \iota + 16\beta\gamma^2 d H^2 \iota + 8 H \beta\sqrt{2 d \iota}\sum_{h=1}^H \sqrt{\sum_{k=1}^K \sigma_{k,h}^2} \nonumber \\
  &\le 8 d H^3 \iota + 16\beta\gamma^2 d H^2 \iota + 8 H \beta\sqrt{2 d H \iota \Vest}, \label{eq:I1-bound}
\end{align}
where the equality uses $\tilde{\beta} = 2 H \beta$ and $\bar{\beta} = \beta$ from \eqref{eq:def-radii}, the first inequality applies \eqref{eq:bonus-sum-bound} of Lemma~\ref{lem:bonus-sum} at each stage $h \in [H]$, and the second inequality follows from the Cauchy--Schwarz inequality and \eqref{eq:def-Vest}. For $I_2$, each summand $D_{k,h}$ in \eqref{eq:def-Dkh} satisfies
\begin{align*}
  D_{k,h}
  &\le \min\cbr{4 d^3 H^2 [\PP_h(V_{k,h+1} - \check{V}_{k,h+1})](s_h^k, a_h^k) + 16 d^3 H^2 \beta\|\bphi(s_h^k, a_h^k)\|_{\bSigma_{k,h}^{-1}}, \, d^3 H^3} \\
  &\le 4 d^3 H^2 [\PP_h(V_{k,h+1} - \check{V}_{k,h+1})](s_h^k, a_h^k) + \min\cbr{16 d^3 H^2 \beta\|\bphi(s_h^k, a_h^k)\|_{\bSigma_{k,h}^{-1}}, \, d^3 H^3},
\end{align*}
where the first inequality follows from \eqref{eq:P1-opt} and \eqref{eq:P1-pess} of Lemma~\ref{lem:double-induction} together with $\bar{\beta} = \beta$, and the second inequality uses $\min\{a + b, c\} \le a + \min\{b, c\}$ for $a := 4 d^3 H^2 [\PP_h(V_{k,h+1} - \check{V}_{k,h+1})](s_h^k, a_h^k) \ge 0$. Summing over $(k,h) \in [K] \times [H]$ yields
\begin{align}
  I_2 &\le 8 d^3 H^2 S_{\ul} + 2\sum_{h=1}^H \sum_{k=1}^K \min\cbr{16 d^3 H^2 \beta\|\bphi(s_h^k, a_h^k)\|_{\bSigma_{k,h}^{-1}}, \, d^3 H^3} \nonumber \\
  &\le 8 d^3 H^2 S_{\ul} + 4 d^4 H^4 \iota + 64 d^4 H^3 \beta\gamma^2 \iota + 32 d^3 H^2 \beta\sqrt{2 d H \iota \Vest} \nonumber \\
  &\le 32 d^3 H^3 \sum_{k=1}^K \sum_{h=1}^H B_{k,h} + 32 d^3 H^5 \ln(H/\delta) + 4 d^4 H^4 \iota + 64 d^4 H^3 \beta\gamma^2 \iota + 32 d^3 H^2 \beta\sqrt{2 d H \iota \Vest} \nonumber \\
  &\le \cO\!\rbr{d^4 H^4 \beta\gamma^2 \iota + d^3 H^5 \ln(H/\delta) + d^{3.5} H^{3.5}\beta\iota^{0.5}\sqrt{\Vest}}, \label{eq:I2-bound}
\end{align}
where the first inequality uses the definition of $S_{\ul}$ in \eqref{eq:def-Sul}, the second inequality applies \eqref{eq:bonus-sum-bound} of Lemma~\ref{lem:bonus-sum} and the Cauchy--Schwarz inequality across $h \in [H]$, the third inequality substitutes \eqref{eq:freed-S-ul} of Lemma~\ref{lem:freedman-value-diff}, and the last inequality substitutes \eqref{eq:sum-Bkh-Vest} of Lemma~\ref{lem:bonus-sum}. Combining \eqref{eq:Vest-master-decomp}, \eqref{eq:I1-bound}, \eqref{eq:I2-bound}, and \eqref{eq:freed-S-pi} (together with \eqref{eq:sum-Bkh-Vest}), we obtain
\[
  \Vest \le \min\cbr{\Varkstar, \Varkpi} + \cO\!\rbr{d^{3.5} H^{3.5}\beta\iota^{0.5}\sqrt{\Vest} + d^7 H^7 \beta \iota^3},
\]
where we used $\gamma^2 = 4 d^3 H^3 \iota^2$ and $\ln(H/\delta) \le \iota_0 \le \beta$. Applying the implication $x \le A + B\sqrt{x} \implies x \le 2 A + B^2$ (for $x, A, B \ge 0$) and substituting $\beta = \cO(\sqrt{d\iota^2\iota_0} + \iota\iota_0)$ establishes \eqref{eq:total-est-var-bound}.
\end{proof}

\begin{proof}[Proof of Theorem~\ref{thm:main}]
Condition on the joint event $\Econc \cap \EFreed$, which holds with probability at least $1 - 3\delta$ by Lemmas~\ref{lem:Econc} and~\ref{lem:freedman-value-diff} and a union bound. We have
\begin{align*}
  \Regret(K)
  &\le 2\sum_{k=1}^K \sum_{h=1}^H B_{k,h} + 4 H^2 \ln(H/\delta) \\
  &\le 8\beta\sqrt{2 d H \iota \Vest} + 16\beta\gamma^2 d H \iota + 16 d H^2 \iota + 4 H^2 \ln(H/\delta) \\
  &\le \cO\!\rbr{\beta\sqrt{d H \iota \cdot \min\cbr{\Varkstar, \Varkpi}} + d^4 H^4 \beta^2 \iota + d^4 H^4 \beta \iota^3} \\
  &\le \cO\!\rbr{\big(d \iota^{1.5}\sqrt{\iota_0} + \sqrt{d}\iota^{1.5}\iota_0\big)\sqrt{H \cdot \min\cbr{\Varkstar, \Varkpi}} + d^5 H^4 \iota^4 \iota_0 + d^4 H^4 \iota^4 \iota_0^2},
\end{align*}
where the first inequality is \eqref{eq:freed-regret} of Lemma~\ref{lem:freedman-value-diff}, the second inequality substitutes \eqref{eq:sum-Bkh-Vest} of Lemma~\ref{lem:bonus-sum}, the third inequality uses $\sqrt{\Vest} \le \sqrt{2\min\{\Varkstar, \Varkpi\}} + \cO(d^{3.5} H^{3.5}\beta\iota^{0.5} + d^{3.5} H^{3.5}\beta^{0.5}\iota^{1.5})$ from \eqref{eq:total-est-var-bound} of Lemma~\ref{lem:total-est-var} and $\beta\gamma^2 d H \iota = 4 d^4 H^4 \beta\iota^3$, and the last inequality substitutes $\beta = \cO(\sqrt{d\iota^2\iota_0} + \iota\iota_0)$ (so that $\beta^2 = \cO(d\iota^2\iota_0 + \iota^2\iota_0^2)$ and $d^{4.5}\iota_0^{0.5} \le d^5\iota_0 + d^4$). Finally, the bound $\cO(d H \iota)$ on the number of value updates follows from Lemma~\ref{lem:switching-count}, and the computational complexity $\tilde{\cO}(d^4 H^3 |\cA| K)$ follows from Remark~\ref{rem:runtime}.
\end{proof}

\section{Trajectory Variance Bounds and Proof of Corollary~\ref{cor:first-order}}
\label{app:corollaries}

To prove the worst-case and first-order regret consequences in Remark~\ref{rem:minimax-adaptivity} and Corollary~\ref{cor:first-order}, we establish high-probability bounds on the realized trajectory variance $\Varkpi$ using a telescoping variance self-bounding argument \citep{chen2021implicit,chen2022improved,foster2024behavior}.

\begin{lemma}[Trajectory Variance Bounds]
\label{lem:traj-var}
Suppose that $r_h(s,a) \ge 0$ for all $(s,a,h) \in \cS \times \cA \times [H]$ and $\sum_{h=1}^H r_h(s_h, a_h) \le B_V$ almost surely for any trajectory. Then for any $\delta \in (0,1)$, the following two bounds hold:
\begin{enumerate}
  \item \textbf{Worst-case trajectory variance bound.} With probability at least $1 - \delta$,\footnote{For $B_V = H$, \eqref{eq:traj-var-worst} gives $\Varkpi \le 4 H^2 K + 16 H^2 \ln(2/\delta)$, shaving a factor of $H$ off the logarithmic term in the $\cO(H^2 K + H^3 \ln(1/\delta))$ total variance lemma \citep[Lemma~C.5]{jin2018q} (which applies an episode-level Azuma--Bernstein inequality to $\sum_{h=1}^H [\VV_h V_{h+1}^{\pi^k}](s_h^k, a_h^k) \in [0, H^3]$), while extending directly to the first-order bound \eqref{eq:traj-var-first-order}.}
  \begin{equation}
    \Varkpi \le 4 B_V^2 K + 16 B_V^2 \ln(2/\delta). \label{eq:traj-var-worst}
  \end{equation}
  \item \textbf{First-order trajectory variance bound.} With probability at least $1 - \delta$,
  \begin{equation}
    \Varkpi \le 6 B_V \sum_{k=1}^K V_1^{\pi^k}(s_1^k) + 32 B_V^2 \ln(4/\delta) \le 6 B_V \Vtotalstar + 32 B_V^2 \ln(4/\delta). \label{eq:traj-var-first-order}
  \end{equation}
\end{enumerate}
\end{lemma}

\begin{proof}
Under the trajectory return assumption, along any realized trajectory we have $0 \le V_h^{\pi^k}(s_h^k) \le B_V$ and $0 \le V_{h+1}^{\pi^k}(s_{h+1}^k) \le B_V$ almost surely for all $(k,h) \in [K] \times [H]$. Using $V_{H+1}^{\pi^k}(s_{H+1}^k) = 0$, we expand $\Varkpi$ via a telescoping sum over $h = 1, \dots, H$:
\begin{align}
  \Varkpi &= \sum_{k=1}^K \sum_{h=1}^H \rbr{[\PP_h (V_{h+1}^{\pi^k})^2](s_h^k, a_h^k) - \big([\PP_h V_{h+1}^{\pi^k}](s_h^k, a_h^k)\big)^2} \nonumber \\
  &= \underbrace{\sum_{k=1}^K \sum_{h=1}^H \rbr{[\PP_h (V_{h+1}^{\pi^k})^2](s_h^k, a_h^k) - \big(V_{h+1}^{\pi^k}(s_{h+1}^k)\big)^2}}_{=: \cT_1} \nonumber \\
  &\quad + \underbrace{\sum_{k=1}^K \sum_{h=1}^H \rbr{\big(V_h^{\pi^k}(s_h^k)\big)^2 - \big([\PP_h V_{h+1}^{\pi^k}](s_h^k, a_h^k)\big)^2}}_{=: \cT_2} - \sum_{k=1}^K \big(V_1^{\pi^k}(s_1^k)\big)^2. \label{eq:Varkpi-telescope}
\end{align}
For the martingale difference sum $\cT_1$, let $X_{k,h} := [\PP_h (V_{h+1}^{\pi^k})^2](s_h^k, a_h^k) - (V_{h+1}^{\pi^k}(s_{h+1}^k))^2$. Since $V_{h+1}^{\pi^k}(s_{h+1}^k) \in [0, B_V]$ almost surely given $\cF_{k,h}$, we have $|X_{k,h}| \le B_V^2$ almost surely, $\EE[X_{k,h} \mid \cF_{k,h}] = 0$, and
\[
  \EE\big[X_{k,h}^2 \mid \cF_{k,h}\big] = \big[\VV_h (V_{h+1}^{\pi^k})^2\big](s_h^k, a_h^k) \le 4 B_V^2 [\VV_h V_{h+1}^{\pi^k}](s_h^k, a_h^k),
\]
where the inequality follows from Lemma~\ref{lem:var-product} applied to $V_{h+1}^{\pi^k}(s_{h+1}^k) \in [0, B_V]$ conditioned on $\cF_{k,h}$. Applying Lemma~\ref{lem:freedman} with $\eta = 1/(8 B_V^2) \in (0, 1/B_V^2]$ and failure probability $\delta/2$ yields that with probability at least $1 - \delta/2$,
\begin{equation}
  \cT_1 \le \eta \sum_{k=1}^K \sum_{h=1}^H 4 B_V^2 [\VV_h V_{h+1}^{\pi^k}](s_h^k, a_h^k) + \frac{\ln(2/\delta)}{\eta} = \frac{1}{2}\Varkpi + 8 B_V^2 \ln(2/\delta). \label{eq:T1-freed-bound}
\end{equation}
For the Bellman-difference sum $\cT_2$, we have
\begin{equation}
  \cT_2 = \sum_{k=1}^K \sum_{h=1}^H \big(V_h^{\pi^k}(s_h^k) + [\PP_h V_{h+1}^{\pi^k}](s_h^k, a_h^k)\big) r_h(s_h^k, a_h^k) \le 2 B_V \sum_{k=1}^K \sum_{h=1}^H r_h(s_h^k, a_h^k), \label{eq:T2-reward-bound}
\end{equation}
where the equality uses the Bellman equation $V_h^{\pi^k}(s_h^k) = r_h(s_h^k, a_h^k) + [\PP_h V_{h+1}^{\pi^k}](s_h^k, a_h^k)$, and the inequality holds since $V_h^{\pi^k}(s_h^k) \le B_V$ and $[\PP_h V_{h+1}^{\pi^k}](s_h^k, a_h^k) \le B_V$ almost surely. Because $\sum_{h=1}^H r_h(s_h^k, a_h^k) \le B_V$ almost surely for every $k \in [K]$, \eqref{eq:T2-reward-bound} deterministically gives $\cT_2 \le 2 B_V^2 K$. Substituting \eqref{eq:T1-freed-bound} and $\cT_2 \le 2 B_V^2 K$ into \eqref{eq:Varkpi-telescope} and rearranging proves \eqref{eq:traj-var-worst}.

To prove \eqref{eq:traj-var-first-order}, let $R_k := \sum_{h=1}^H r_h(s_h^k, a_h^k) \in [0, B_V]$ denote the realized return in episode $k$. Since $R_k$ is $\cF_{k+1,1}$-measurable with $\EE[R_k \mid \cF_{k,1}] = V_1^{\pi^k}(s_1^k) \le V_1^*(s_1^k)$, applying Corollary~\ref{cor:nonneg-mult} to $\{R_k\}_{k=1}^K$ with range $B_V$ and failure probability $\delta/2$ gives, with probability at least $1 - \delta/2$,
\[
  \sum_{k=1}^K \sum_{h=1}^H r_h(s_h^k, a_h^k) = \sum_{k=1}^K R_k \le \frac{3}{2}\sum_{k=1}^K V_1^{\pi^k}(s_1^k) + 4 B_V \ln(4/\delta).
\]
Substituting this bound into \eqref{eq:T2-reward-bound}, combining with \eqref{eq:T1-freed-bound} via a union bound, and rearranging \eqref{eq:Varkpi-telescope} establishes \eqref{eq:traj-var-first-order}.
\end{proof}

\begin{proof}[Proof of Corollary~\ref{cor:first-order}]
In the setting of Theorem~\ref{thm:main}, we have $B_V = H$, so \eqref{eq:traj-var-first-order} of Lemma~\ref{lem:traj-var} holds with probability at least $1 - \delta$:
\[
  \min\cbr{\Varkstar, \Varkpi} \le \Varkpi \le 6 H \Vtotalstar + 32 H^2 \ln(4/\delta).
\]
Substituting this upper bound into \eqref{eq:main-regret-bound} of Theorem~\ref{thm:main} and taking a union bound completes the proof.
\end{proof}

\section{Comparison with Time-Homogeneous Horizon-Free RL}
\label{app:homog-comparison}

While the main text focuses on episodic time-inhomogeneous MDPs with per-step reward $r_h \in [0,1]$ (Table~\ref{tab:comparison}), a parallel line of research studies \emph{time-homogeneous} episodic MDPs ($\PP_1 = \dots = \PP_H = \PP$ and $r_1 = \dots = r_H = r$) under the bounded total reward normalization $\sum_{h=1}^H r(s_h, a_h) \le 1$, aiming to achieve \emph{horizon-free} regret bounds with no polynomial dependence on $H$. Table~\ref{tab:comparison-homog} summarizes existing results for time-homogeneous linear mixture MDPs and linear MDPs in this regime.

\begin{table}[h!]
\centering
\caption{Comparison of regret bounds for episodic time-homogeneous RL with linear function approximation under bounded total reward $\sum_{h=1}^H r_h \le 1$. Regret upper bounds are shown up to universal constants and polylogarithmic factors in $d, H, K, 1/\delta$. Here $\Varkstar = \sum_{k=1}^K \sum_{h=1}^H [\VV V_{h+1}^*](s_h^k, a_h^k)$ denotes the realized cumulative trajectory variance of the optimal value function.}
\label{tab:comparison-homog}
\vspace{4pt}
\begingroup
\footnotesize
\setlength{\tabcolsep}{6pt}
\renewcommand{\arraystretch}{1.35}
\begin{tabular}{l c c c}
\toprule
\textbf{Algorithm} & \textbf{Regret Bound} & \textbf{Deterministic Regime} & \textbf{Efficient?} \\
\midrule
\multicolumn{4}{l}{\textit{Time-Homogeneous Linear Mixture MDPs}} \\
\algcell{\texttt{VARLin}}{\citep{zhang2021improved}}
& $d^{4.5}\sqrt{K}+d^9$
& $d^{4.5}\sqrt{K}+d^9$
& $\times$ \\
\addlinespace[2pt]
\algcell{\texttt{VARLin2}}{\citep{kim2022improved}}
& $d\sqrt{K}+d^2$
& $d\sqrt{K}+d^2$
& $\times$ \\
\addlinespace[2pt]
\algcell{\hfucrlvtrp{}}{\citep{zhou2022computationally}}
& $d\sqrt{K}+d^2$
& $d\sqrt{K}+d^2$
& $\checkmark$ \\
\addlinespace[2pt]
\algcell{\ucrlave{}}{\citep{zhao2023variance}}
& $d\sqrt{\Varkstar}+d^2$
& $d^2$
& $\checkmark$ \\
\midrule
\multicolumn{4}{l}{\textit{Time-Homogeneous Linear MDPs}} \\
\algcell{Algorithm~1}{\citep{zhang2024horizon}}
& $d^{5.5}\sqrt{K}+d^{6.5}$
& $d^{5.5}\sqrt{K}+d^{6.5}$
& $\times$ \\
\bottomrule
\end{tabular}
\endgroup
\end{table}

As shown in Table~\ref{tab:comparison-homog}, in time-homogeneous linear mixture MDPs, the transition kernel $\PP(s'\mid s,a) = \inner{\bphi(s'\mid s,a)}{\btheta^*}$ is governed by a single $d$-dimensional parameter vector $\btheta^*$ shared across all stages $h \in [H]$, allowing samples from all $H$ stages to be pooled into a single regression problem; exploiting this property, \citet{zhou2022computationally} (\hfucrlvtrp{}) and \citet{zhao2023variance} (\ucrlave{}) achieved computationally efficient horizon-free worst-case and variance-dependent regret bounds, respectively. In time-homogeneous linear MDPs, by contrast, even though the signed transition measures $\btheta(\cdot)$ are stage-independent, the optimal value functions $\{V_{h+1}^*\}_{h=1}^H$ remain time-inhomogeneous across stages, inducing $H$ distinct Bellman backup parameter vectors $\wb(V_{h+1}^*) = \int_{\cS} V_{h+1}^*(s')\dd\btheta(s')$ in general. Consequently, pooling state-action features across stages faces a covariance mismatch across different stage variances $[\VV V_{h+1}^*]$. Recently, \citet{zhang2024horizon} overcame this statistical barrier via a computationally inefficient model-elimination algorithm that maintains function-indexed covariance matrices over an $\epsilon$-net of value functions and groups stages by bounding the total variation of $\{V_h^*\}_{h=1}^H$, establishing a worst-case horizon-free regret bound of $\tilde{\cO}(d^{5.5}\sqrt{K} + d^{6.5})$. Whether one can achieve horizon-free \emph{variance-dependent} regret in time-homogeneous linear MDPs, and whether horizon-free regret can be attained by a computationally efficient algorithm, remain open problems.

\section{Auxiliary Lemmas}

In this appendix, we collect the concentration inequalities and technical facts from prior work used throughout our proofs.

\subsection{Martingale Concentration Inequalities}

We first recall the time-uniform Freedman inequality for bounded martingale difference sequences \citep{freedman1975tail,beygelzimer2011contextual}, stated below in the form of \citet[Lemma~D.2]{foster2024behavior}.

\begin{lemma}[Time-Uniform Freedman Inequality, {\citealt[Lemma~D.2]{foster2024behavior}}]
\label{lem:freedman}
Let $\{M_t\}_{t=1}^T$ be a real-valued martingale difference sequence adapted to a filtration $\{\cG_t\}_{t=0}^T$ such that $|M_t| \le B$ almost surely for all $t \in [T]$. Then for any fixed $\eta \in (0, 1/B]$ and any $\delta' \in (0,1)$, with probability at least $1 - \delta'$, for all $T' \in [T]$:
\[
  \sum_{t=1}^{T'} M_t \le \eta \sum_{t=1}^{T'} \EE\big[M_t^2 \mid \cG_{t-1}\big] + \frac{\ln(1/\delta')}{\eta}.
\]
\end{lemma}

\begin{corollary}[Multiplicative Concentration, {\citealt[Lemma~D.3]{foster2024behavior}}]
\label{cor:nonneg-mult}
Let $\{X_t\}_{t=1}^T$ be a sequence of random variables adapted to a filtration $\{\cG_t\}_{t=0}^T$ such that $0 \le X_t \le B$ almost surely for all $t \in [T]$. Then for any $\delta' \in (0,1)$, with probability at least $1 - \delta'$, for all $T' \in [T]$:
\[
  \sum_{t=1}^{T'} X_t \le \frac{3}{2}\sum_{t=1}^{T'} \EE[X_t \mid \cG_{t-1}] + 4 B \ln(2/\delta').
\]
\end{corollary}

\begin{lemma}[Variance-Aware Self-Normalized Concentration Inequality, {\citealt[Theorem~4.3]{zhou2022computationally}}]
\label{lem:bernstein-sn}
Let $\{\cG_i\}_{i=1}^\infty$ be a filtration, and let $\{\xb_i, \eta_i\}_{i \ge 1}$ be a stochastic process such that $\xb_i \in \RR^d$ is $\cG_i$-measurable and $\eta_i \in \RR$ is $\cG_{i+1}$-measurable. Suppose that for constants $\sigma, R, L_x > 0$,
\[
  \EE[\eta_i \mid \cG_i] = 0, \qquad \EE[\eta_i^2 \mid \cG_i] \le \sigma^2, \qquad |\eta_i| \le R, \qquad \|\xb_i\|_2 \le L_x.
\]
Let $\Zb_k = \lambda\Ib + \sum_{i=1}^{k-1} \xb_i\xb_i^\top$ for $\lambda > 0$. Then for any $\delta' \in (0,1)$ and any discretization parameter $\epsilon_{\mathrm{c}} \in (0,R]$, with probability at least $1 - \delta'$, for all $k \in [K]$:
\[
  \nbr{\sum_{i=1}^{k-1} \xb_i \eta_i}_{\Zb_k^{-1}} \le 12\sqrt{\sigma^2 d \ln\!\rbr{1 + \frac{K L_x^2}{d\lambda}}\iota_{\mathrm{c}}(\delta')} + 24\iota_{\mathrm{c}}(\delta') \max_{1 \le i \le k-1} |\eta_i|\min\cbr{1, \|\xb_i\|_{\Zb_i^{-1}}} + 6\iota_{\mathrm{c}}(\delta')\epsilon_{\mathrm{c}},
\]
where $\iota_{\mathrm{c}}(\delta') := \ln(32(\ln(R/\epsilon_{\mathrm{c}}) + 1)K^2/\delta')$.
\end{lemma}

\subsection{Miscellaneous Bounds}

We next record a variance bound for products of bounded random variables from \citet[Lemma~30]{chen2021implicit} (see also \citealt[Lemma~34]{chen2022improved}).

\begin{lemma}[{\citealt[Lemma~30]{chen2021implicit}}]
\label{lem:var-product}
For any two bounded real-valued random variables $X$ and $Y$, we have
\[
  \Var(X Y) \le 2 \Var(X)\|Y\|_\infty^2 + 2\big(\EE[X]\big)^2 \Var(Y).
\]
In particular, taking $Y = X$ with $\|X\|_\infty \le B$ gives $\Var(X^2) \le 4 B^2 \Var(X)$.
\end{lemma}

Applying Lemma~\ref{lem:var-product} (see also \citealt[Lemma~G.8]{foster2024behavior}) to the conditional distribution $\PP_h(\cdot \mid s,a)$ immediately yields the following inequality for squared value functions.

\begin{corollary}[Conditional Variance Bound for Squared Value Functions]
\label{cor:var-squared}
For any measurable function $V: \cS \to [0, B_V]$ and any state-action-stage triplet $(s,a,h) \in \cS \times \cA \times [H]$, we have
\[
  [\VV_h V^2](s,a) \le 4 B_V^2 [\VV_h V](s,a).
\]
\end{corollary}

\begin{proof}
Fix $(s,a,h)$ and let $X = V(s') \in [0, B_V]$ where $s' \sim \PP_h(\cdot \mid s,a)$. Since $\|X\|_\infty \le B_V$, Lemma~\ref{lem:var-product} with $Y = X$ yields $[\VV_h V^2](s,a) = \Var(X^2) \le 4 B_V^2 \Var(X) = 4 B_V^2 [\VV_h V](s,a)$.
\end{proof}

\begin{lemma}[Elliptical Potential Lemma, {\citealt[Lemma~11]{abbasi2011improved}}]
\label{lem:elliptical-potential}
Let $\{\xb_k\}_{k=1}^K \subset \RR^d$ be a sequence of vectors satisfying $\|\xb_k\|_2 \le L_x$ for all $k \in [K]$, and define $\Zb_k = \lambda\Ib + \sum_{i=1}^{k-1} \xb_i\xb_i^\top$ for $\lambda > 0$. Then
\[
  \sum_{k=1}^K \min\cbr{1, \, \|\xb_k\|_{\Zb_k^{-1}}^2} \le 2 d \ln\!\rbr{1 + \frac{K L_x^2}{d\lambda}}.
\]
\end{lemma}

\begin{lemma}[Determinant Ratio Inequality, {\citealt[Lemma~12]{abbasi2011improved}}]
\label{lem:det-ratio}
Suppose $\Ab, \Bb \in \RR^{d\times d}$ are two positive definite matrices satisfying $\Ab \succeq \Bb$. Then for any $\xb \in \RR^d$, we have $\|\xb\|_{\Ab} \le \|\xb\|_{\Bb} \cdot \sqrt{\det(\Ab)/\det(\Bb)}$.
\end{lemma}

\begin{lemma}[Covering Number of Euclidean Balls, {\citealt[Lemma~D.5]{jin2020provably}}]
\label{lem:euclidean-covering}
For any $\epsilon > 0$ and $R > 0$, the $\epsilon$-covering number of the Euclidean ball $\{\xb \in \RR^d : \|\xb\|_2 \le R\}$ under the Euclidean norm $\|\cdot\|_2$ is upper-bounded by $(1 + 2 R / \epsilon)^d$.
\end{lemma}

\bibliographystyle{ims}
\bibliography{ref}

\end{document}